\documentclass{article}
\usepackage[T1]{fontenc}
\usepackage[utf8]{inputenc}
\usepackage{times}
\usepackage[authoryear,round]{natbib}
\setcitestyle{citesep={;},aysep={,},yysep={;}}
\usepackage{fancyhdr}

\makeatletter
\def\normalsize{\@setsize\normalsize{11pt}\xpt\@xpt}
\def\small{\@setsize\small{10pt}\ixpt\@ixpt}
\def\footnotesize{\@setsize\footnotesize{10pt}\ixpt\@ixpt}
\def\scriptsize{\@setsize\scriptsize{8pt}\viipt\@viipt}
\def\tiny{\@setsize\tiny{7pt}\vipt\@vipt}
\def\large{\@setsize\large{14pt}\xiipt\@xiipt}
\def\Large{\@setsize\Large{16pt}\xivpt\@xivpt}
\def\LARGE{\@setsize\LARGE{20pt}\xviipt\@xviipt}
\def\section{\@startsection{section}{1}{\z@}{-2ex plus -.5ex minus -.2ex}{1.5ex plus .3ex minus .2ex}{\large\scshape\raggedright}}
\def\subsection{\@startsection{subsection}{2}{\z@}{-1.8ex plus -.5ex minus -.2ex}{.8ex plus .2ex}{\normalsize\scshape\raggedright}}
\def\subsubsection{\@startsection{subsubsection}{3}{\z@}{-1.5ex plus -.5ex minus -.2ex}{.5ex plus .2ex}{\normalsize\scshape\raggedright}}
\def\paragraph{\@startsection{paragraph}{4}{\z@}{1.5ex plus .5ex minus .2ex}{-1em}{\normalsize\bfseries}}
\renewcommand{\maketitle}{%
  \par\begingroup
  {\LARGE\rmfamily\raggedright\@title\par}
  \vskip .15in
  {\normalsize\raggedright\@author\par}
  \vskip .20in
  \endgroup
  \begingroup
  \renewcommand{\thefootnote}{\fnsymbol{footnote}}%
  \footnotetext[1]{Equal contribution (joint first authors).}%
  \footnotetext[2]{Corresponding author.}%
  \endgroup
  \setcounter{footnote}{0}%
}
\makeatother
\renewenvironment{abstract}{\vskip.075in\centerline{\large\scshape Abstract}\vspace{.5ex}\begin{quote}}{\par\end{quote}\vskip1ex}
\skip\footins9pt plus4pt minus2pt
\renewcommand{\footnoterule}{\kern-3pt\hrule width12pc\kern2.6pt}
\belowdisplayskip\abovedisplayskip
\usepackage{amsmath,amssymb,bm}
\usepackage{booktabs}
\usepackage{array}
\usepackage{longtable}
\usepackage{pdflscape}
\usepackage{colortbl}
\usepackage{graphicx}

\newcommand{\paperfigure}[2][]{%
  \IfFileExists{#2}{\includegraphics[#1]{#2}}{%
    \PackageWarningNoLine{pod-rope}{External figure unavailable: #2}%
    \fbox{\parbox[c][0.16\textheight][c]{0.94\linewidth}{%
      \centering\small\textbf{External figure unavailable in this preview}\par
      \medskip\texttt{\detokenize{#2}}\par\medskip
      Compile the complete source package to display the original figure.}}%
  }%
}
\usepackage{afterpage}

\usepackage{multirow}
\usepackage{xcolor}
\definecolor{oursblue}{RGB}{20,92,166}
\definecolor{oursrow}{RGB}{238,246,252}
\usepackage{microtype}
\usepackage{xspace}
\usepackage{tikz}
\usetikzlibrary{arrows.meta,positioning}

\newcommand{\papertablestyle}{\normalsize\rmfamily\setlength{\tabcolsep}{4pt}\renewcommand{\arraystretch}{1}\setlength{\heavyrulewidth}{0.08em}\setlength{\lightrulewidth}{0.05em}\setlength{\cmidrulewidth}{0.03em}}

\usepackage[hidelinks]{hyperref}
\usepackage{url}
\newcommand{\method}{\textnormal{POD-RoPE}\xspace}
\newcommand{\qkmethod}{\textnormal{POD-RoPE (Q/K-only)}\xspace}
\newcommand{\pointmethod}{\textnormal{Physical Point (Q/K-only)}\xspace}

\title{POD-RoPE: Rotary Position Encoding over Physical Observation Distributions for Embodied AI}
\author{%
\begingroup\fontsize{8}{10}\selectfont\bfseries
\makebox[\textwidth][s]{%
Xiaodi Huang\textsuperscript{1,2,3,*}\hfill
Zilin Wang\textsuperscript{2,*}\hfill
Jinpeng Liu\textsuperscript{3,4}\hfill
Pengchuan Xiao\textsuperscript{3}\hfill
Zhenlei Shao\textsuperscript{3}\hfill
Zhang Zhang\textsuperscript{1,\textdagger}}
\endgroup\\[0.35em]
\parbox[t]{\textwidth}{\fontsize{9}{11}\selectfont\raggedright
\mbox{\textsuperscript{1}Institute of Automation, Chinese Academy of Sciences}\hspace{1em}%
\mbox{\textsuperscript{2}University of Chinese Academy of Sciences}\\
\mbox{\textsuperscript{3}Sharpa}\hspace{1em}%
\mbox{\textsuperscript{4}National University of Singapore}}
}
\date{}
\hypersetup{pdftitle={POD-RoPE: Rotary Position Encoding over Physical Observation Distributions for Embodied AI},pdfsubject={Preprint},pdfauthor={Xiaodi Huang, Zilin Wang, Jinpeng Liu, Pengchuan Xiao, Zhenlei Shao, Zhang Zhang}}

\begin{document}
\raggedbottom
\setlength{\parskip}{4pt}
\setlength{\textfloatsep}{10pt plus 2pt minus 1pt}
\setlength{\floatsep}{8pt plus 2pt minus 1pt}
\setlength{\intextsep}{8pt plus 2pt minus 1pt}
\setlength{\abovecaptionskip}{5pt}
\maketitle

\afterpage{
\begin{figure}[!t]
  \centering
  \paperfigure[width=0.96\textwidth]{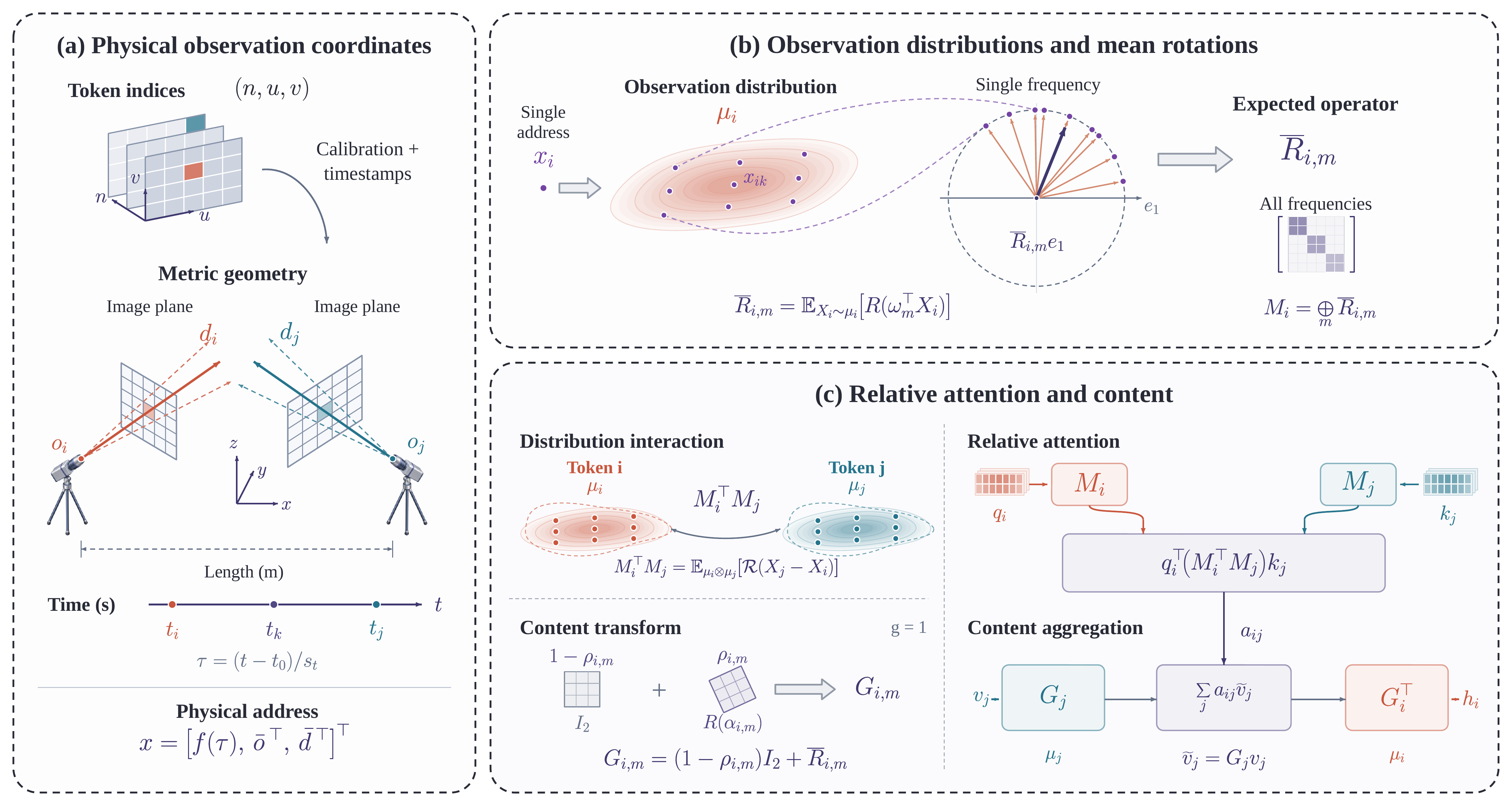}
  \caption{\textbf{\method: physical observation distributions for attention and content aggregation.}
  (a) Calibration and timestamps map token indices to physical addresses.
  (b) Averaging address-dependent rotations under $\mu_i$ gives $M_i$; vectors illustrate one frequency acting on $e_1$.
  (c) For independent address samples, $M_i^\top M_j$ equals the expected relative rotation and guides Q/K matching. The full integration also transforms values and aggregated outputs using $G_{i,m}=(1-\rho_{i,m})I_2+\overline R_{i,m}$ ($g=1$), where $\rho_{i,m}$ measures rotation coherence. Distributions and rotations are schematic; rays indicate viewing directions.}
  \label{fig:overview}
\end{figure}
}
\begin{abstract}
Embodied perception requires positional representations that capture both physical geometry and the spatial and temporal extent of observations. We introduce \textbf{POD-RoPE} (Physical Observation Distribution RoPE), which represents token observation footprints as distributions over camera centers, viewing directions, and sampling times. Under independent sampling, expected rotations encode relative relationships between distributions; when each distribution collapses to a single address, the construction recovers pointwise RoPE. The same rotation statistics support Q/K matching and coherence-dependent V/O transformations, allowing observation geometry to guide both information selection and content aggregation. Per-token pre-aggregation avoids increasing sequence length or processing pairs of observation nodes. We evaluate \method across four robot policy architectures, simulation and real-world manipulation, and complementary reconstruction, depth-estimation, and novel-view-synthesis tasks. Integrating POD-RoPE into Fast-WAM improves mean success from 91.83\% to 92.88\% across all 50 RoboTwin 2.0 tasks, and from 32.5\% to 52.5\% on two real-robot tasks on Sharpa North. Controlled reconstruction ablations support the contributions of distribution encoding and V/O transformations. In H20 model-block benchmarks, \method achieves median forward-plus-backward speedups of $1.91\times$ over P-RoPE and $5.34\times$ over RayRoPE, with a median latency overhead of 17.4\% over Standard RoPE. These results support physical observation distributions as an effective and computationally efficient basis for geometry-aware attention.
\end{abstract}

\section{Introduction}
\label{sec:intro}

Perception and interaction in the physical world require embodied systems to establish physically meaningful spatial and temporal relationships between observations. Standard RoPE~\citep{su2021roformer}, however, typically defines relative positions using sequence indices or spatiotemporal grid coordinates: identical pixel offsets can correspond to different viewing angles, and identical frame-index offsets can represent different elapsed times. Positional encoding therefore needs to extend beyond discrete index differences to represent observation geometry and actual spatial and temporal scales.

Geometric positional encodings such as P-RoPE and RayRoPE incorporate camera and ray information into relative relationship modeling~\citep{li2025cameras,wu2026rayrope}, but incur substantially higher latency than Standard RoPE in our model-block benchmarks. Meanwhile, an image token aggregates information over a pixel region, and a temporally compressed video token combines observations from multiple times. A single center address cannot fully describe these observation footprints. Our central view is that positional encoding should represent not only \emph{where and when a token is observed, but also the spatial and temporal extent of its observations}. We investigate how to encode these physical observation distributions and their relative relationships at low computational overhead.

We propose \method, which maps camera centers, viewing directions, and sampling times into a shared physical address space and represents each token's observation footprint as an address distribution. Averaging address-dependent rotations yields a token-level positional operator; under independent sampling, the pairwise Q/K operator equals the expected rotation over the relative-address distribution (Eq.~\eqref{eq:expected_relative_rotation}). When each distribution concentrates at a single address, the construction recovers pointwise RoPE at the corresponding addresses. The full method reuses these statistics for V/O transformations, allowing geometry to guide both matching and aggregation. Per-token pre-aggregation avoids extra tokens and pairwise processing of observation nodes, limiting computational overhead.

We evaluate \method across four robot policy architectures, three visual-geometry tasks, and two real-robot tasks on Sharpa North. Across all 50 RoboTwin 2.0 tasks, full integration into Fast-WAM raises mean success from 91.83\% to 92.88\%; on Sharpa North, it raises mean success from 32.5\% to 52.5\%. Visual-geometry experiments show task-dependent gains, and controlled ablations isolate the two design components. In H20 microbenchmarks, the full method achieves median forward-plus-backward speedups of $1.91\times$ over P-RoPE and $5.34\times$ over RayRoPE, with 17.4\% median overhead over Standard RoPE. Our contributions are:
\begin{itemize}
\setlength{\itemsep}{2pt}
\setlength{\parsep}{0pt}
\setlength{\topsep}{4pt}
  \item \textbf{Physical observation distribution encoding.} We represent token footprints as distributions over camera centers, viewing directions, and sampling times. Expected rotations encode relative-address interactions and recover pointwise RoPE for singleton distributions.
  \item \textbf{Efficient matching and aggregation.} Per-token rotation statistics are shared by Q/K matching and coherence-dependent V/O transformations, avoiding additional tokens and pairwise observation-node processing.
  \item \textbf{Cross-domain validation.} Tests across four robot policy architectures and three visual-geometry tasks show improved aggregate manipulation success and task-dependent geometry gains. Ablations support both components; model-block benchmarks show lower latency than P-RoPE and RayRoPE.
\end{itemize}

\section{Related Work}

\paragraph{Rotary positional encoding and geometry-aware attention.}
RoPE applies position-dependent rotations to queries and keys so that positional interactions in attention depend on their relative displacement~\citep{su2021roformer}. For multi-view vision, GTA applies group transformations associated with camera geometry to attention matching and value-feature aggregation~\citep{miyato2024gta}. P-RoPE (denoted PRoPE in the original paper) further unifies intrinsic and extrinsic camera parameters through projective transformations, incorporating complete view-frustum relationships into relative positional representations~\citep{li2025cameras}. RayRoPE constructs projective ray coordinates in the query-camera frame and predicts depth and its uncertainty. It uses expected RoPE to encode position distributions along rays and applies the transformations to Q/K/V/O~\citep{wu2026rayrope}. \method addresses the observation footprints induced by tokenization. It jointly encodes camera centers, unit viewing directions, and actual time in a shared reference frame, averaging over the address distributions covered by pixels and source frames. Rotation statistics can be pre-aggregated per token and reused across attention interactions. The full method further uses rotation coherence to modulate V/O content transformations with an identity branch.

\paragraph{Observation footprints and spatiotemporal positional representations.}
Extending positional encoding from points to regions allows representations to retain observation scale. Mip-NeRF approximates conical frustums with Gaussian distributions and averages sinusoidal positional features to obtain scale-aware integrated positional encoding~\citep{barron2021mipnerf}. RayRoPE applies expected rotations to uncertainty intervals of predicted depth~\citep{wu2026rayrope}. Meanwhile, DiT performs diffusion modeling on image-latent patches~\citep{peebles2023dit}, and Wan compresses spatial and temporal dimensions through a causal video autoencoder~\citep{wan2025}, so one visual token corresponds to multiple pixels and potentially multiple source frames. We map these nominal footprints into physical observation distributions, using camera calibration and timestamps to determine their addresses and the phase and magnitude of average rotations to retain the observation composition. This applies regional integration to relative matching and content aggregation between compressed visual tokens, with one construction shared by static multi-view and video observations.

\paragraph{Vision-action policies and world-action models.}
ACT predicts action chunks~\citep{zhao2023act}, RDT uses a multimodal diffusion Transformer~\citep{liu2024rdt}, and $\pi_0$ combines a pretrained vision-language model with flow-matching action generation~\citep{black2024pi0}. Fast-WAM jointly models video and actions during training but omits explicit future-video generation at inference~\citep{yuan2026fastwam}. We study observation geometry and footprints as positional inputs across these policy architectures; Section~\ref{sec:embodied_control} specifies their integrations.

\paragraph{Embodied data and visual geometry evaluation.}
We evaluate manipulation on RoboTwin 2.0~\citep{mu2024robotwin,chen2025robotwin2} and Sharpa North, and reconstruction using RoboTwin, AgiBot World~\citep{bu2025agibot}, and collected Sharpa North observations. Complementary geometry tasks use UniMatch~\citep{xu2023unimatch} with the DeMoN depth protocol~\citep{ummenhofer2017demon}, and LVSM~\citep{jin2025lvsm} for novel-view synthesis on RealEstate10K~\citep{zhou2018stereo}.

\section{Method}
\label{sec:method}

\method incorporates camera geometry and actual sampling time into positional encoding through physical observation addresses, observation distributions, and their expected rotations. We next describe the address and distribution constructions, relative Q/K relationships, and efficient computation and V/O content aggregation through shared rotation statistics (Figure~\ref{fig:overview}).

\subsection{From Discrete Indices to Physical Observation Addresses}

Discrete indices describe the arrangement of tokens in a tensor but do not directly express camera baselines, viewing directions, or actual time intervals. We use these observed quantities as positional inputs to RoPE. For pixel $p$ of camera $c$ at time $t$, let $\widetilde p$ denote homogeneous pixel coordinates, $K_c(t)$ the intrinsics, and $(Q_c(t),b_c(t))$ the extrinsics mapping camera coordinates to a shared reference frame. For undistorted pinhole images with matching intrinsics, the camera center and unit viewing direction are
\begin{equation}
  \begin{aligned}
    o(t,c)&=b_c(t),\\[3pt]
    d(t,p,c)&=Q_c(t)\frac{K_c(t)^{-1}\widetilde p}
      {\|K_c(t)^{-1}\widetilde p\|_2}.
  \end{aligned}
  \label{eq:physical_ray}
\end{equation}
Normalizing actual time, camera centers, and directions by specified scales gives the physical observation address
\begin{equation}
  x=g(t,p,c)
  =\bigl[f(\tau(t)),\,\bar o(t,c)^\top,\,\bar d(t,p,c)^\top\bigr]^\top.
  \label{eq:physical_address}
\end{equation}
where $\tau=(t-t_0)/s_t$, $\bar o=o/s_o$, and $\bar d=s_dd$. The parameters $s_t$ and $s_o$ specify time and length scales, respectively, while $s_d$ adjusts the directional component. The temporal mapping $f$ controls how phase changes with time; our experiments use $f(\tau)=2\tanh(\tau/2)$. Static observations omit the time component. This address describes where, in which direction, and when an observation is made, without assuming the unknown surface depth along the ray. Robot experiments use camera-to-robot-root extrinsics. Appendix~\ref{sec:physical_coordinates} details the calibration chain, intrinsic updates, and actual-time mapping.

\subsection{From a Single Address to an Observation Distribution}

A token typically aggregates a pixel region, and a video token may also integrate multiple times. Even with physical addresses, encoding only the center therefore omits the observation footprint: tokens with the same central viewing direction or time can cover different fields of view or temporal spans. We represent the physical addresses covered by token $i$ and their normalized weights as a probability distribution $\mu_i$, with $X_i\sim\mu_i$. It describes the composition of observations within the token, rather than uncertainty about an unknown scene position. Appendix~\ref{sec:support_measure_definition} gives the construction.

To incorporate these observations into positional encoding, we first define the rotation for a single address and then average it over the observation distribution. For head dimension $d_h=2M$, choose frequency vectors $\omega_m\in\mathbb R^D$ and define
\begin{equation}
  \mathcal R(x)=\bigoplus_{m=1}^{M}R(\omega_m^\top x),
  \label{eq:point_group_representation}
\end{equation}
where $R(\theta)$ is a 2D rotation matrix and $\bigoplus$ places the rotation blocks along the diagonal. The token's positional operator is
\begin{equation}
  M_i=\mathbb E_{X_i\sim\mu_i}[\mathcal R(X_i)].
  \label{eq:support_operator}
\end{equation}
This construction allows all observations covered by a token to determine its positional encoding jointly. When the distribution concentrates at one address $x_i$, $M_i=\mathcal R(x_i)$, recovering pointwise RoPE at that physical address.

\subsection{Preserving Relative Relationships between Distributions}

After averaging rotations, the key question is whether RoPE's relative-position property is preserved. Pointwise rotations satisfy
\begin{equation}
  \mathcal R(x)^\top\mathcal R(y)=\mathcal R(y-x).
  \label{eq:point_relative}
\end{equation}
Thus, independently encoding two addresses yields the rotation of their relative address through pairwise interaction. Distribution encoding preserves this structure:
\begin{equation}
  M_i^\top M_j
  =\mathbb E[\mathcal R(X_j-X_i)].
  \label{eq:expected_relative_rotation}
\end{equation}
Here, $X_i\sim\mu_i$ and $X_j\sim\mu_j$ are sampled independently, representing all pairwise combinations of the two observation sets under their respective weights. The equality follows directly from the relative-position property of pointwise rotations and linearity of expectation; Appendix~\ref{sec:relative_proof} provides the proof.

The interaction between two positional operators is therefore determined by the distribution of relative observation addresses, extending relative relationships between points to those between observation sets. Relative addresses contain the mapped time difference $f(\tau_j)-f(\tau_i)$, camera-center displacement, and viewing-direction difference. Let $q_i,k_j$ denote queries and keys after projection and any model-required normalization, but before the positional operator. The attention logit is
\begin{equation}
  \ell_{ij}=\frac{(M_iq_i)^\top(M_jk_j)}{\sqrt{d_h}}.
  \label{eq:distribution_attention_logit}
\end{equation}
Equation~\eqref{eq:expected_relative_rotation} thus directly constrains pairwise positional interactions in attention. Each model applies its required Q/K normalization before the positional operator; Appendix~\ref{sec:attention_order} specifies the order.

\subsection{Efficient Computation and Distribution-Guided Content Aggregation}

\paragraph{Computing expected rotations.}
For each token and frequency, the mean cosine and sine suffice to construct the positional operator:
\begin{equation}
  \begin{aligned}
    c_{i,m}&=\mathbb E[\cos(\omega_m^\top X_i)],\qquad
    s_{i,m}=\mathbb E[\sin(\omega_m^\top X_i)],\\
    \overline R_{i,m}&=
    \begin{bmatrix}c_{i,m}&-s_{i,m}\\s_{i,m}&c_{i,m}\end{bmatrix},
    \qquad M_i=\bigoplus_m\overline R_{i,m}.
  \end{aligned}
  \label{eq:expected_rotation}
\end{equation}
These statistics are reused across all attention interactions without processing pairs of observation nodes. Encoding retains Standard RoPE's computation over 2D feature blocks. Appendix~\ref{sec:quadrature_implementation} provides numerical quadrature and complexity details.

\paragraph{Constructing content transformations from rotation coherence.}
Average rotations generally combine rotation and contraction. For $\rho_{i,m}>0$, we have $\overline R_{i,m}=\rho_{i,m}R(\alpha_{i,m})$, where $\rho_{i,m}=\sqrt{c_{i,m}^2+s_{i,m}^2}\leq1$ measures rotation coherence and $\alpha_{i,m}$ is the mean direction. Directly transforming content with $M_i$ would suppress value features when rotations cancel, so we introduce an identity branch:
\begin{equation}
  G_{i,m}=(1-g\rho_{i,m})I_2+g\overline R_{i,m},\qquad
  G_i=\bigoplus_mG_{i,m},\qquad g\in[0,1].
  \label{eq:content_operator_pair}
\end{equation}
This mixes the identity and the rotation in the mean direction with weights $1-g\rho_{i,m}$ and $g\rho_{i,m}$. Greater coherence assigns more weight to the rotation branch. At $\rho_{i,m}=0$, the expression directly gives $G_{i,m}=I_2$, preserving content without defining a mean direction. Appendices~\ref{sec:center_information_proof} and~\ref{sec:operator_properties} analyze coherence under fixed sampling templates and the transformation properties.

\paragraph{Distribution-guided content aggregation.}
Let $a_{ij}=\operatorname{softmax}_j(\ell_{ij})$. We transform values before weighted aggregation and apply an output transformation afterward:
\begin{equation}
  h_i=G_i^\top\sum_j a_{ij}G_jv_j.
  \label{eq:operator_attention}
\end{equation}
Here, $a_{ij}$ weights each token's contribution, while $G_i^\top G_j$ transforms its content features. Setting $g=0$ yields Q/K-only integration; the full method fixes $g=1$ without adding learnable gating parameters. Appendix Table~\ref{tab:method_configurations} lists model-specific integrations.

\section{Experiments}
\label{sec:experiments}
\label{sec:results}
Robot experiments evaluate the complete policy integrations; Table~\ref{tab:component_ablation} tests the contributions of distribution encoding and V/O transformations. Visual geometry and model-block benchmarks assess quality and efficiency. Appendices~\ref{sec:experiment_setup_details} and~\ref{sec:real_robot_setup} provide complete training and evaluation protocols.

\subsection{End-to-End Embodied Manipulation}
\label{sec:embodied_control}
\paragraph{Simulation evaluation setup.}
We use two training regimes on RoboTwin 2.0~\citep{chen2025robotwin2}. Multitask training uses Fast-WAM~\citep{yuan2026fastwam} as the baseline and jointly trains a shared policy on 2,500 Clean demonstrations and 25,000 randomized-scene demonstrations from 50 tasks. Single-task training learns a separate policy for each task; we adapt $\pi_0$, RDT, and ACT on 15 tasks with 50 Clean demonstrations per task. Both regimes evaluate 100 episodes per task under each of the Clean and Randomized conditions and report task-averaged success rates. Randomized conditions vary clutter, lighting, textures, and table height. Both comparisons follow the training recipe and evaluation settings of their corresponding baselines. All policies use Q/K+V/O integration. Appendices~\ref{sec:fastwam_details} and~\ref{sec:robotwin_single_details} provide detailed configurations.

\paragraph{Joint multitask training.}
\begin{table}[t]
\centering
\papertablestyle

\caption{Fast-WAM success rates (\%) on 50 RoboTwin 2.0 tasks. Baseline~\citep{yuan2026fastwam} is Fast-WAM; Ours integrates \method into Fast-WAM. Both use identical training and evaluation settings. Light blue shading identifies our method; the best average results are bold.}
\label{tab:fastwam50}
\begin{tabular}{@{}lrr>{\columncolor{oursrow}}r>{\columncolor{oursrow}}r@{}}
\toprule
 & \multicolumn{2}{c}{Baseline} & \multicolumn{2}{c}{Ours}\\
\cmidrule(lr){2-3}\cmidrule(lr){4-5}
Task & Clean & Rand. & Clean & Rand.\\
\midrule
adjust\_bottle & 100 & 100 & 100 & 100\\
beat\_block\_hammer & 99 & 97 & 100 & 98\\
blocks\_ranking\_rgb & 100 & 100 & 100 & 100\\
blocks\_ranking\_size & 94 & 98 & 95 & 94\\
click\_alarmclock & 100 & 100 & 100 & 100\\
$\vdots$ & $\vdots$ & $\vdots$ & $\vdots$ & $\vdots$\\
\midrule
Average (50 tasks) & 91.88 & 91.78 & \textbf{93.12} & \textbf{92.64}\\
$\Delta$ (pp) & -- & -- & +1.24 & +0.86\\
\bottomrule
\end{tabular}
\end{table}

Table~\ref{tab:fastwam50} shows gains under both Clean and Randomized conditions, raising mean success across the two conditions from 91.83\% to 92.88\%. Integrating \method into Fast-WAM improves task-averaged success by 1.24 and 0.86 percentage points under Clean and Randomized conditions, respectively. The table lists the first five tasks in their original order; Appendix Table~\ref{tab:fastwam50_tasks} reports all 50, including task-level regressions.

\paragraph{Single-task training across architectures.}
\begin{table}[!t]
\centering
\papertablestyle
\setlength{\tabcolsep}{0.8pt}

\caption{Success rates (\%) after single-task fine-tuning on 15 RoboTwin 2.0 tasks across different model architectures. Each model is compared under its corresponding baseline's training and evaluation configuration. Baseline~\citep{chen2025robotwin2} is the original policy; Ours integrates full \method. Rand. denotes Randomized conditions; bold marks the best result per task and condition across the six configurations.}\label{tab:robotwin_single_all_tasks}
\label{tab:robotwin15}
\begin{tabular}{@{}lcc>{\columncolor{oursrow}}c>{\columncolor{oursrow}}ccc>{\columncolor{oursrow}}c>{\columncolor{oursrow}}ccc>{\columncolor{oursrow}}c>{\columncolor{oursrow}}c@{}}
\toprule
\multirow{3}{*}{Task} & \multicolumn{4}{c}{$\pi_0$} & \multicolumn{4}{c}{RDT} & \multicolumn{4}{c}{ACT}\\
\cmidrule(lr){2-5}\cmidrule(lr){6-9}\cmidrule(lr){10-13}
 & \multicolumn{2}{c}{Baseline} & \multicolumn{2}{c}{Ours} & \multicolumn{2}{c}{Baseline} & \multicolumn{2}{c}{Ours} & \multicolumn{2}{c}{Baseline} & \multicolumn{2}{c}{Ours}\\
\cmidrule(lr){2-3}\cmidrule(lr){4-5}\cmidrule(lr){6-7}\cmidrule(lr){8-9}\cmidrule(lr){10-11}\cmidrule(lr){12-13}
 & Clean & Rand. & Clean & Rand. & Clean & Rand. & Clean & Rand. & Clean & Rand. & Clean & Rand.\\
\midrule
adjust bottle & 90 & 56 & \textbf{100} & \textbf{100} & 81 & 75 & 97 & 88 & 97 & 23 & 99 & 26\\
beat block hammer & 43 & 21 & 51 & 33 & 77 & \textbf{37} & \textbf{89} & 33 & 56 & 3 & 77 & 0\\
blocks ranking rgb & 19 & 5 & \textbf{26} & \textbf{22} & 3 & 0 & 9 & 1 & 1 & 0 & 4 & 0\\
click bell & 44 & 3 & \textbf{96} & \textbf{24} & 80 & 9 & 83 & 0 & 58 & 3 & 92 & 8\\
handover block & 45 & 8 & 53 & 15 & 45 & 14 & \textbf{67} & \textbf{16} & 42 & 0 & 28 & 0\\
hanging mug & 11 & 3 & 15 & 10 & \textbf{23} & \textbf{16} & 16 & 6 & 7 & 0 & 7 & 0\\
lift pot & 84 & 36 & \textbf{95} & \textbf{47} & 72 & 9 & 88 & 22 & 88 & 0 & 90 & 0\\
move can pot & 58 & 21 & 49 & \textbf{31} & 25 & 12 & 29 & 18 & 22 & 4 & \textbf{60} & 5\\
open laptop & \textbf{85} & 46 & 84 & 40 & 59 & 32 & 72 & \textbf{50} & 56 & 0 & 68 & 0\\
open microwave & 80 & 50 & \textbf{100} & \textbf{63} & 37 & 20 & 90 & 29 & 86 & 0 & 85 & 0\\
pick dual bottles & \textbf{57} & 12 & 49 & \textbf{22} & 42 & 13 & 55 & 11 & 31 & 0 & 46 & 0\\
place object basket & 16 & 2 & \textbf{72} & \textbf{57} & 33 & 17 & 45 & 11 & 15 & 0 & 20 & 0\\
press stapler & 62 & 29 & \textbf{80} & \textbf{36} & 41 & 24 & 43 & 26 & 31 & 6 & 49 & 7\\
stack blocks three & 17 & 0 & \textbf{31} & \textbf{14} & 2 & 0 & 7 & 1 & 0 & 0 & 20 & 0\\
stack bowls two & \textbf{91} & 41 & 82 & \textbf{55} & 76 & 30 & 83 & 34 & 82 & 0 & 85 & 0\\
\midrule
Mean & 53.47 & 22.20 & \textbf{65.53} & \textbf{37.93} & 46.40 & 20.53 & 58.20 & 23.07 & 44.80 & 2.60 & 55.33 & 3.07\\
$\Delta$ (pp) & -- & -- & +12.07 & +15.73 & -- & -- & +11.80 & +2.53 & -- & -- & +10.53 & +0.47\\
\bottomrule
\end{tabular}
\end{table}

After integrating \method, all three single-task policies achieve higher average success rates (Table~\ref{tab:robotwin_single_all_tasks}). $\pi_0$ benefits under both conditions, whereas RDT and ACT improve primarily under Clean conditions; ACT's Randomized success rate remains 3.07\%. The gains therefore vary substantially across architectures and evaluation conditions.

\paragraph{Real-robot manipulation on Sharpa North.}
We further evaluate bottle opening and straw insertion on the Sharpa North bimanual robot to evaluate the complete \method integration in real manipulation. Both methods train jointly on the two tasks, use observations from the head's left-eye camera and both wrist cameras, and are evaluated for 20 trials per task at the 20,000-step checkpoint. The real-robot version enables both Q/K distribution encoding and V/O content transformations in the video and action branches. Appendix~\ref{sec:real_robot_setup} gives the training settings.
\begin{table}[t]
\centering
\papertablestyle
\caption{Real-robot task results on Sharpa North. Both methods use the same architecture; the baseline uses standard RoPE. Both use the 20,000-step checkpoint and are evaluated for 20 trials per task; success rates are in parentheses. $\Delta$ denotes the success-rate difference in percentage points.}
\label{tab:real_robot}
\begin{tabular}{@{}lr>{\columncolor{oursrow}}rr@{}}
\toprule
Task & Fast-WAM & Fast-WAM + \method & $\Delta$ (pp)\\
\midrule
Bottle opening & 7/20 (35\%) & 12/20 (60\%) & +25.0\\
Straw insertion & 6/20 (30\%) & 9/20 (45\%) & +15.0\\
\midrule
Average (2 tasks) & 32.5\% & \textbf{52.5\%} & +20.0\\
\bottomrule
\end{tabular}
\end{table}

\begin{figure}[t]
  \centering
  \paperfigure[width=\textwidth]{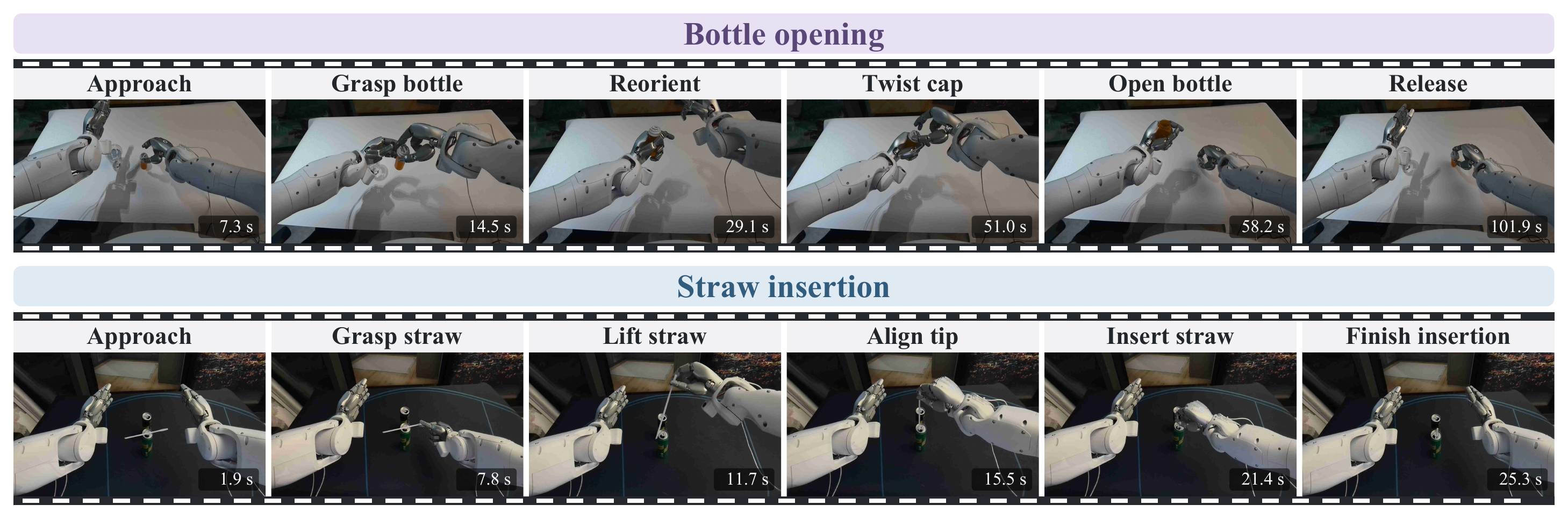}
  \caption{Bottle-opening and straw-insertion examples on Sharpa North. Each row shows six moments from one successful training demonstration, ordered from left to right; times are relative to the start of that recording. The images illustrate the task process. Table~\ref{tab:real_robot} reports the real-robot evaluation results of the two policies.}
  \label{fig:real_robot_tasks}
\end{figure}

Both real-robot tasks improve, raising mean success from 32.5\% to 52.5\% (Table~\ref{tab:real_robot}). These tasks test bimanual coordination, rotation, and small-object alignment. Figure~\ref{fig:real_robot_tasks} illustrates their demonstration sequences.

\subsection{Further Evaluation on Visual Geometry Tasks}
\label{sec:visual_geometry}
\begin{table}[!t]
\centering\papertablestyle
\caption{Cross-view reconstruction and $h16$ conditional temporal extrapolation with different positional encodings on three datasets. At $h16$, the target occurs 16 time steps after the reference observations, with each step spanning four original synchronized sampling intervals: approximately 2.15, 4.27, and 2.13 seconds on Sharpa North, RoboTwin, and AGIBot, respectively. Bold marks the best result within each dataset and evaluation setting.}
\label{tab:embodied_h16}

\setlength{\tabcolsep}{2.5pt}
\begin{tabular}{@{}llrrrrrrrr@{}}
\toprule
& & \multicolumn{4}{c}{Cross-view Reconstruction} & \multicolumn{4}{c}{$h16$ Conditional Reconstruction}\\
\cmidrule(lr){3-6}\cmidrule(lr){7-10}
Dataset & Method & MSE$\downarrow$ & PSNR$\uparrow$ & SSIM$\uparrow$ & LPIPS$\downarrow$ & MSE$\downarrow$ & PSNR$\uparrow$ & SSIM$\uparrow$ & LPIPS$\downarrow$\\
\midrule
Sharpa North & Standard RoPE & 0.1931 & 16.264 & 0.5682 & 0.5668 & 0.3078 & 15.204 & 0.5364 & 0.5937\\
Sharpa North & P-RoPE & 0.1129 & 18.825 & 0.6476 & 0.2728 & 0.2114 & 17.160 & 0.5931 & 0.3678\\
\rowcolor{oursrow}Sharpa North & \method & \textbf{0.1022} & \textbf{19.376} & \textbf{0.6779} & \textbf{0.2371} & \textbf{0.1853} & \textbf{17.808} & \textbf{0.6279} & \textbf{0.3164}\\
\midrule
RoboTwin & Standard RoPE & 0.1378 & 22.064 & 0.7474 & 0.3921 & 0.4716 & 17.733 & 0.6917 & 0.5085\\
RoboTwin & P-RoPE & 0.1335 & 22.459 & 0.7517 & 0.3451 & 0.3961 & 18.104 & 0.6964 & 0.4720\\
\rowcolor{oursrow}RoboTwin & \method & \textbf{0.1130} & \textbf{23.302} & \textbf{0.7643} & \textbf{0.3235} & \textbf{0.3631} & \textbf{18.571} & \textbf{0.7036} & \textbf{0.4552}\\
\midrule
AGIBot & Standard RoPE & 0.2841 & 18.164 & 0.5768 & 0.5249 & 0.3377 & 18.285 & 0.5869 & 0.5279\\
AGIBot & P-RoPE & 0.1862 & 20.622 & 0.6354 & \textbf{0.3308} & 0.2477 & 20.223 & 0.6314 & \textbf{0.3533}\\
\rowcolor{oursrow}AGIBot & \method & \textbf{0.1806} & \textbf{20.998} & \textbf{0.6392} & 0.3319 & \textbf{0.2387} & \textbf{20.709} & \textbf{0.6363} & 0.3564\\

\bottomrule
\end{tabular}
\end{table}

\paragraph{Synchronized cross-view reconstruction.}
\begin{samepage}
We reconstruct a third view from two simultaneous camera observations on Sharpa North, RoboTwin, and AGIBot, using shared splits and trajectory/task-balanced evaluation. Table~\ref{tab:embodied_h16} compares Standard RoPE, P-RoPE, and a fixed \method configuration. Relative to P-RoPE, \method improves all four metrics on Sharpa North and RoboTwin, and improves PSNR but slightly worsens LPIPS on AGIBot. Spatial reconstruction gains are thus dataset-dependent. Appendix Figure~\ref{fig:embodied_reconstruction} shows examples from identical reference observations.
\par
\end{samepage}

\paragraph{Conditional temporal extrapolation.}
We also fix the current reference observations and supply the future target's camera geometry and time. Under matched settings, \method improves future-target PSNR over P-RoPE on all three datasets (Table~\ref{tab:embodied_h16}), supporting physical spatiotemporal encoding for conditional temporal reconstruction.

Appendix~\ref{sec:temporal_horizon_analysis} provides complete curves and analysis across temporal horizons, and Appendix~\ref{sec:embodied_input_changes} reports experiments with input changes.

\begin{table}[!t]
\centering
\papertablestyle
\setlength{\tabcolsep}{2.4pt}
\caption{Depth estimation with different positional encodings on UniMatch/DeMoN. All methods use the same backbone, training, and evaluation settings; bold marks the best results.}

\begin{tabular}{@{}llrrrrrrr@{}}
\toprule
Dataset & Method & AbsRel$\downarrow$ & SqRel$\downarrow$ & RMSE$\downarrow$ & RMSE-log$\downarrow$ & $\delta_1\uparrow$ & $\delta_2\uparrow$ & $\delta_3\uparrow$\\
\midrule
\multirow{5}{*}{RGBD}
& Standard RoPE & 0.1753 & 0.2523 & 0.9085 & 0.2716 & 0.7273 & 0.8741 & 0.9407\\
& P-RoPE & 0.1231 & 0.2129 & 0.6839 & 0.2115 & 0.8523 & \textbf{0.9264} & 0.9463\\
& RayRoPE & \textbf{0.1221} & 0.2303 & \textbf{0.6668} & 0.2071 & \textbf{0.8563} & 0.9244 & 0.9429\\
\rowcolor{oursrow}& \qkmethod & 0.1307 & \textbf{0.1982} & 0.7041 & 0.2051 & 0.8133 & 0.9194 & \textbf{0.9491}\\
\rowcolor{oursrow}& \method & 0.1243 & 0.2094 & 0.6753 & \textbf{0.2033} & 0.8403 & 0.9227 & 0.9454\\
\midrule
\multirow{5}{*}{Scenes11}
& Standard RoPE & 0.1185 & 0.2060 & 0.9395 & 0.2104 & 0.8660 & 0.9422 & 0.9700\\
& P-RoPE & 0.0681 & 0.0939 & 0.6149 & 0.1332 & 0.9455 & 0.9816 & 0.9908\\
& RayRoPE & 0.0597 & 0.0810 & 0.5390 & 0.1216 & \textbf{0.9603} & \textbf{0.9863} & 0.9918\\
\rowcolor{oursrow}& \qkmethod & 0.0696 & 0.0954 & 0.6002 & 0.1337 & 0.9490 & 0.9829 & 0.9899\\
\rowcolor{oursrow}& \method & \textbf{0.0594} & \textbf{0.0739} & \textbf{0.5306} & \textbf{0.1215} & \textbf{0.9603} & 0.9850 & \textbf{0.9923}\\
\bottomrule
\end{tabular}
\label{tab:unimatch_demon}
\end{table}

\paragraph{Depth estimation.}
We compare five positional encodings on UniMatch/DeMoN under identical backbone, training, and evaluation settings (Table~\ref{tab:unimatch_demon}). Full \method improves 12 of 14 metrics over \qkmethod and 10 over P-RoPE. It leads on five Scenes11 metrics and ties RayRoPE on $\delta_1$, whereas RayRoPE performs better on four RGBD metrics. Section~\ref{sec:efficiency} separately evaluates model-block cost, and Appendix~\ref{sec:experiment_setup_details} gives the setup.

\paragraph{Novel-view synthesis.}
Under matched 80k training on RE10K/LVSM, full \method improves all three quality metrics over \qkmethod, including a 0.398 dB PSNR gain. Its PSNR exceeds Pl\"ucker Raymap and GTA, while RayRoPE remains best on all three metrics and leads \method by 0.832 dB. Figure~\ref{fig:efficiency_comparison}(b) shows the quality--latency trade-off; Appendix Table~\ref{tab:re10k} reports all metrics.

\subsection{Ablation Studies}
\label{sec:ablation}
Physical point Q/K raises PSNR from 18.138 to 21.832 dB over Standard RoPE (Table~\ref{tab:component_ablation}), supporting physical addresses with their frequency construction.

With coordinates and frequencies fixed, distribution Q/K improves PSNR/SSIM/LPIPS by $+0.534$/$+0.0259$/$-0.0373$ over point Q/K. With V/O enabled, distributions improve all three metrics over points, including 0.744 dB in PSNR. Both comparisons support encoding token footprints beyond single addresses.

V/O improves all metrics with both encodings, raising PSNR by 0.340 dB for points and 0.550 dB for distributions. Direct-average V/O instead underperforms distribution Q/K-only by 0.373 dB and worsens SSIM/LPIPS. While $\overline R=\rho R(\alpha)$ contracts features at low coherence, adding $(1-\rho)I$ makes the operator approach identity as coherence vanishes. The full method gains 0.923 dB over direct averaging, supporting this identity branch (Appendix~\ref{sec:component_ablation_setup}).
\begin{table}[!t]
\centering\papertablestyle
\caption{Component ablation on Sharpa North head-to-wrist reconstruction under matched settings. Direct-average V/O keeps distribution Q/K fixed. Bold marks the best results.}
\label{tab:component_ablation}
\begin{tabular}{@{}lrrr@{}}
\toprule
Configuration & PSNR$\uparrow$ & SSIM$\uparrow$ & LPIPS$\downarrow$\\
\midrule
Standard RoPE & 18.138 & 0.6576 & 0.6014\\
\pointmethod & 21.832 & 0.7206 & 0.3194\\
Physical Point (Q/K+V/O) & 22.172 & 0.7331 & 0.2997\\
\rowcolor{oursrow}\qkmethod & 22.366 & 0.7465 & 0.2821\\
\rowcolor{oursrow}Distribution + direct-average V/O & 21.993 & 0.7255 & 0.3341\\
\rowcolor{oursrow}\method & \textbf{22.916} & \textbf{0.7524} & \textbf{0.2655}\\
\bottomrule
\end{tabular}
\end{table}

\subsection{Computational Efficiency}
\label{sec:efficiency}
\begin{figure}[!t]
  \centering
  \begin{minipage}[b]{0.54\linewidth}
    \centering
    \resizebox{\linewidth}{!}{\begin{tikzpicture}[x=0.60cm,y=0.85cm,font=\sffamily]
    \definecolor{classiclight}{RGB}{237,232,224}
    \definecolor{classicdark}{RGB}{151,146,138}
    \definecolor{supportlight}{RGB}{146,205,236}
    \definecolor{supportdark}{RGB}{52,132,184}
    \definecolor{operatorlight}{RGB}{92,151,207}
    \definecolor{operatordark}{RGB}{20,72,132}
    \definecolor{plight}{RGB}{246,177,104}
    \definecolor{pdark}{RGB}{218,104,20}
    \definecolor{raylight}{RGB}{218,112,88}
    \definecolor{raydark}{RGB}{154,43,30}
    \definecolor{gtalight}{RGB}{181,145,207}
    \definecolor{gtadark}{RGB}{102,58,139}

    \draw[line width=0.6pt] (0.90,0.85) -- (0.90,5.15);
    \draw[line width=0.6pt] (0.90,0.85) -- (12.00,0.85);
    \foreach \y/\lab in {0.85/0,1.85/25,2.85/50,3.85/75,4.85/100} {
      \draw[gray!25,dashed,line width=0.35pt] (0.90,\y) -- (12.00,\y);
      \node[anchor=east,font=\scriptsize] at (0.75,\y) {\lab};
    }
    \node[rotate=90,font=\scriptsize\bfseries] at (0.05,3.00) {Latency (ms, $\downarrow$)};

    \filldraw[fill=classiclight,draw=black,line width=0.45pt] (1.20,0.85) rectangle (2.20,1.066);
    \filldraw[fill=classicdark,draw=black,line width=0.45pt] (1.20,1.066) rectangle (2.20,1.313);
    \filldraw[fill=supportlight,draw=black,line width=0.45pt] (3.00,0.85) rectangle (4.00,1.033);
    \filldraw[fill=supportdark,draw=black,line width=0.45pt] (3.00,1.033) rectangle (4.00,1.319);
    \filldraw[fill=operatorlight,draw=black,line width=0.45pt] (4.80,0.85) rectangle (5.80,1.111);
    \filldraw[fill=operatordark,draw=black,line width=0.45pt] (4.80,1.111) rectangle (5.80,1.544);
    \filldraw[fill=plight,draw=black,line width=0.45pt] (6.60,0.85) rectangle (7.60,1.402);
    \filldraw[fill=pdark,draw=black,line width=0.45pt] (6.60,1.402) rectangle (7.60,2.393);
    \filldraw[fill=raylight,draw=black,line width=0.45pt] (8.40,0.85) rectangle (9.40,2.423);
    \filldraw[fill=raydark,draw=black,line width=0.45pt] (8.40,2.423) rectangle (9.40,4.888);
    \filldraw[fill=gtalight,draw=black,line width=0.45pt] (10.20,0.85) rectangle (11.20,1.396);
    \filldraw[fill=gtadark,draw=black,line width=0.45pt] (10.20,1.396) rectangle (11.20,2.452);

    \node[font=\tiny\bfseries] at (1.70,0.958) {5.39};
    \node[font=\tiny\bfseries] at (3.50,0.946) {4.58};
    \node[font=\tiny\bfseries] at (5.30,0.981) {6.54};
    \node[font=\tiny\bfseries] at (7.10,1.126) {13.81};
    \node[font=\scriptsize\bfseries] at (8.90,1.637) {39.32};
    \node[font=\tiny\bfseries] at (10.70,1.123) {13.66};
    \node[font=\scriptsize\bfseries,text=classicdark] at (1.70,1.46) {11.58};
    \node[font=\scriptsize\bfseries,text=supportdark] at (3.50,1.47) {11.74};
    \node[font=\scriptsize\bfseries,text=operatordark] at (5.30,1.69) {17.34};
    \node[font=\scriptsize\bfseries,text=pdark!80!black] at (7.10,2.54) {38.57};
    \node[font=\scriptsize\bfseries,text=raydark] at (8.90,5.03) {100.94};
    \node[font=\scriptsize\bfseries,text=gtadark] at (10.70,2.60) {40.06};

    \foreach \x/\name in {1.70/Standard RoPE,3.50/{\shortstack{POD-RoPE\\(Q/K-only)}},5.30/POD-RoPE,7.10/P-RoPE,8.90/RayRoPE,10.70/GTA} {
      \node[rotate=28,anchor=east,font=\scriptsize] at (\x,0.66) {\name};
    }

    \filldraw[fill=black!12,draw=black,line width=0.35pt] (3.55,5.34) rectangle (3.82,5.58);
    \node[anchor=west,font=\scriptsize] at (3.92,5.46) {Forward};
    \filldraw[fill=black!55,draw=black,line width=0.35pt] (6.20,5.34) rectangle (6.47,5.58);
    \node[anchor=west,font=\scriptsize] at (6.57,5.46) {Backward only};
  \end{tikzpicture}
}
    \par\vspace{1mm}
    {\small (a) Model-block latency}
  \end{minipage}\hfill
  \begin{minipage}[b]{0.44\linewidth}
    \centering
    \resizebox{\linewidth}{!}{\begin{tikzpicture}[y=1.15cm,font=\sffamily]
    \definecolor{pluckergray}{RGB}{120,126,134}
    \definecolor{supportblue}{RGB}{102,178,224}
    \definecolor{operatorblue}{RGB}{20,92,166}
    \definecolor{gtapurple}{RGB}{126,87,194}
    \definecolor{baselineorange}{RGB}{235,132,42}
    \definecolor{baselinered}{RGB}{173,55,36}

    \draw[line width=0.55pt] (0.70,0.75) -- (0.70,4.15);
    \draw[line width=0.55pt] (0.70,0.75) -- (6.15,0.75);
    \foreach \x/\lab in {0.70/0,1.78/20,2.86/40,3.94/60,5.02/80,6.10/100} {
      \draw[gray!25,dashed,line width=0.35pt] (\x,0.75) -- (\x,4.15);
      \node[font=\scriptsize] at (\x,0.47) {\lab};
    }
    \foreach \y/\lab in {0.75/20,1.43/22,2.11/24,2.79/26,3.47/28,4.15/30} {
      \draw[gray!25,dashed,line width=0.35pt] (0.70,\y) -- (6.15,\y);
      \node[anchor=east,font=\scriptsize] at (0.57,\y) {\lab};
    }
    \node[font=\scriptsize\bfseries] at (3.42,0.05) {Forward+backward latency (ms, $\downarrow$)};
    \node[rotate=90,font=\scriptsize\bfseries] at (-0.08,2.45) {RE10K PSNR (dB, $\uparrow$)};

    \filldraw[fill=supportblue,draw=black,line width=0.55pt] (1.334,2.391) circle[radius=1.2mm];
    \filldraw[fill=pluckergray,draw=black,line width=0.55pt] (1.401,1.920) circle[radius=1.2mm];
    \filldraw[fill=operatorblue,draw=black,line width=0.55pt] (1.637,2.526) circle[radius=1.4mm];
    \filldraw[fill=baselineorange,draw=black,line width=0.55pt] (2.783,2.281) circle[radius=1.2mm];
    \filldraw[fill=gtapurple,draw=black,line width=0.55pt] (2.739,2.215) circle[radius=1.2mm];
    \filldraw[fill=baselinered,draw=black,line width=0.55pt] (6.151,2.809) circle[radius=1.2mm];

    \draw[gray!60,line width=0.35pt] (1.334,2.391) -- (1.65,3.30);
    \node[anchor=south,font=\scriptsize\bfseries,text=supportblue!65!black] at (1.65,3.32) {\shortstack{POD-RoPE\\(Q/K-only)}};
    \draw[gray!60,line width=0.35pt] (1.637,2.526) -- (3.05,3.50);
    \node[anchor=south west,font=\scriptsize\bfseries,text=operatorblue] at (3.05,3.49) {POD-RoPE};
    \draw[gray!60,line width=0.35pt] (1.401,1.920) -- (1.66,1.37);
    \node[anchor=north west,font=\scriptsize,text=pluckergray!80!black] at (1.66,1.39) {Pl\"ucker Raymap};
    \draw[gray!60,line width=0.35pt] (2.783,2.281) -- (3.34,2.82);
    \node[anchor=south west,font=\scriptsize,text=baselineorange!70!black] at (3.34,2.81) {P-RoPE};
    \draw[gray!60,line width=0.35pt] (2.739,2.215) -- (3.45,1.70);
    \node[anchor=north west,font=\scriptsize,text=gtapurple] at (3.45,1.72) {GTA};
    \node[anchor=south east,font=\scriptsize,text=baselinered] at (6.02,2.87) {RayRoPE};
  \end{tikzpicture}
}
    \par\vspace{1mm}
    {\small (b) RE10K quality--latency trade-off}
  \end{minipage}
  \caption{Computation cost and reconstruction quality on an H20.
  (a) Six-layer reconstruction-block latency at $256\times320$, B1, T1, and 960 tokens. Light and dark segments show forward and backward-only time; totals give forward+backward time, using medians over 30 measurements.
  (b) RE10K PSNR versus forward+backward latency of the six-layer LVSM block at $256\times320$, B1, three views, and 960 tokens. RayRoPE takes 100.942 ms and is shown at the 100-ms boundary.
  The panels use separate model-block benchmarks, not end-to-end timings; their absolute latencies are not pooled. Full protocols and values are in Appendix~\ref{sec:efficiency_details} and Table~\ref{tab:re10k}.}
  \label{fig:efficiency_comparison}
\end{figure}
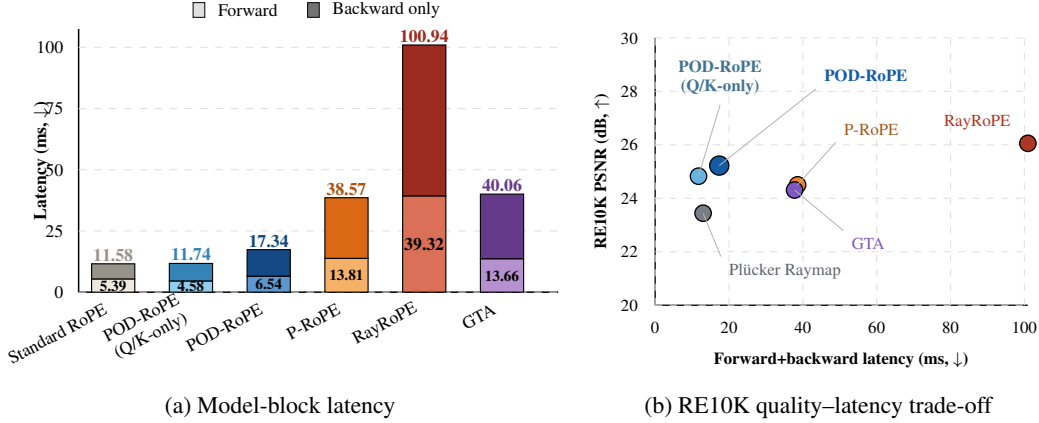

\begin{samepage}
In matched model-block microbenchmarks, full \method takes $1.281\times$/$1.174\times$ the Standard RoPE latency for forward/forward-plus-backward passes; \qkmethod takes $1.073\times$/$1.034\times$. Full \method achieves speedups of $1.81\times$/$1.91\times$ over P-RoPE across seven configurations and $5.69\times$/$5.34\times$ over RayRoPE across five supported configurations, using median per-configuration ratios. Figure~\ref{fig:efficiency_comparison} shows model-block cost and the separate RE10K quality--latency trade-off; Appendix~\ref{sec:efficiency_details} provides complete timings.
\par
\end{samepage}

\section{Conclusion}

\method represents token footprints as distributions of physical observation addresses. Expected rotations preserve relative Q/K interactions, while coherence-dependent V/O transformations reuse these statistics to guide content aggregation. The same construction handles static and temporal observations without increasing sequence length. Complete policy integrations improve aggregate success rates in closed-loop experiments, including Sharpa North manipulation; controlled reconstruction ablations support the contributions of distribution encoding and V/O. Visual geometry evaluations reveal task-dependent gains and quality--cost trade-offs, with low additional model-block latency relative to Standard RoPE.


\begingroup
\normalfont\normalsize
\renewcommand{\bibfont}{\normalfont\normalsize}
\urlstyle{rm}

\endgroup

\appendix
\setlength{\parskip}{4pt}
\section{Method Derivations and Implementation Details}
\label{sec:method_properties}

\subsection{Observation Distributions and the Relative-Position Property}
\label{sec:support_measure_definition}
\label{sec:relative_proof}

\paragraph{Observation distributions.}
The observation distribution $\mu_i$ consists of the physical addresses covered by a token and their normalized weights. For $X_i\sim\mu_i$, $\mathbb E[F(X_i)]$ denotes averaging with observation weights; it is computed using finitely many nodes in practice.

\paragraph{Proof of the relative-position property.}
Two-dimensional rotations satisfy $R(\alpha)^\top R(\beta)=R(\beta-\alpha)$, so Eq.~\eqref{eq:point_group_representation} satisfies $\mathcal R(x)^\top\mathcal R(y)=\mathcal R(y-x)$. Sampling $X_i\sim\mu_i$ and $X_j\sim\mu_j$ independently according to the observation weights of the two tokens gives
\begin{equation}
  \begin{aligned}
    M_i^\top M_j
      &=\mathbb E[\mathcal R(X_i)]^\top
        \mathbb E[\mathcal R(X_j)]\\
      &=\mathbb E[\mathcal R(X_i)^\top\mathcal R(X_j)]\\
      &=\mathbb E[\mathcal R(X_j-X_i)].
  \end{aligned}
\end{equation}
The second step uses independent sampling, and the third uses the relative-position property of pointwise rotations. This proves the Q/K relative-position identity in the main text. For finite nodes, the pairwise weights are $w_{ik}w_{jl}$. Self-interaction of a token also uses two independent samples.

\paragraph{Two immediate consequences.}
When all observations of a token concentrate at address $x_i$, $M_i=\mathcal R(x_i)$, recovering pointwise RoPE with the same addresses and frequencies. If a common offset $a$ is added to every address, $(X_j+a)-(X_i+a)=X_j-X_i$, so the pairwise Q/K positional kernel remains unchanged.

\subsection{Observation Footprints and Rotation Magnitude}
\label{sec:center_information_proof}

\paragraph{An alternative expression for average rotations.}
\label{sec:fourier_interpretation}
The mean cosine and sine can be combined into a complex number:
\begin{equation}
  \Phi_i(\omega_m)
  =c_{i,m}+\mathrm i s_{i,m}
  =\mathbb E[e^{\mathrm i\omega_m^\top X_i}].
  \label{eq:characteristic_function}
\end{equation}
This is the characteristic function of the observation distribution under the positive-exponent convention. It jointly expresses the direction and magnitude of the average rotation at the selected frequency.

\paragraph{Identical centers with different footprints.}
Consider a uniform distribution over one-dimensional physical addresses, $X\sim\mathrm{Unif}[c-a,c+a]$, where $c$ is the center and $a>0$ is the half-width of the footprint. Its average rotation is
\begin{equation}
  \Phi(\omega)
    =\frac{1}{2a}\int_{c-a}^{c+a}e^{\mathrm i\omega x}\,dx
    =e^{\mathrm i\omega c}\frac{\sin(\omega a)}{\omega a}.
  \label{eq:uniform_range_example}
\end{equation}
Center encoding retains only $e^{\mathrm i\omega c}$, while distribution encoding also retains the footprint factor. For $\omega a=\pi/6$ and $\pi/2$, the two distributions have the same center and mean direction but magnitudes of approximately $0.955$ and $0.637$, respectively. Distribution encoding can thus distinguish footprints that the center phase cannot express.

\paragraph{Meaning of the magnitude.}
Each pointwise rotation corresponds to a unit-length vector, whose mean satisfies
\begin{equation}
  \rho_{i,m}=|\Phi_i(\omega_m)|
  \leq\mathbb E[|e^{\mathrm i\omega_m^\top X_i}|]=1.
  \label{eq:fourier_magnitude_main}
\end{equation}
The magnitude equals one if and only if the observation rotations have the same direction at that frequency, allowing phases to differ by integer multiples of $2\pi$. For $\rho_{i,m}>0$,
\begin{equation}
  \overline R_{i,m}
  =\rho_{i,m}R(\arg\Phi_i(\omega_m)).
  \label{eq:circular_mean_factorization}
\end{equation}
The magnitude therefore describes rotation coherence at that frequency. Both widths in the example above lie within $0<\omega a<\pi$; for more general footprints and frequencies, the magnitude should be computed from the actual phase distribution.

\paragraph{Why a fixed sampling template can still yield varying magnitudes.}
Let the fixed pixel nodes be $p_i+\delta_k$ and the address mapping be $g_i$, with $x_{ik}=g_i(p_i+\delta_k)$ and $x_i=g_i(p_i)$. Since the center phase has unit modulus,
\begin{equation}
  \rho_{i,m}=\left|\sum_k w_k e^{\mathrm i\omega_m^\top(x_{ik}-x_i)}\right|.
\end{equation}
The magnitude is determined by physical address offsets. Using the mapping's local Jacobian $J_i$, $x_{ik}-x_i\approx J_i\delta_k$, so a fixed pixel template can still produce different magnitudes as intrinsics, viewing location, or camera orientation change. If the physical offsets are also identical, the magnitude is fixed. The camera-center term shared by the same camera at the same frame does not independently change the magnitude. The V/O ablation compares direct average rotations with and without the coherence-dependent identity branch; it does not separately compare retaining magnitude against phase-only transport.

\subsection{Properties of Distribution-Guided Content Aggregation}
\label{sec:operator_properties}

\paragraph{Norm bound on the transformation.}
Equation~\eqref{eq:content_operator_pair} forms a convex combination of the identity and a rotation. Using $\|\overline R_{i,m}\|_2=\rho_{i,m}$, for $g,\rho_{i,m}\in[0,1]$ we have
\begin{equation}
  \|G_{i,m}\|_2
    \leq(1-g\rho_{i,m})+g\|\overline R_{i,m}\|_2=1.
  \label{eq:operator_bound}
\end{equation}
Thus $\|G_i\|_2\leq1$, so the content transformation itself cannot amplify the feature norm. When $\rho_{i,m}=0$, the block returns to the identity.

\paragraph{Relationship to Q/K relativity.}
Omitting the frequency subscript and writing $\beta_i=1-g\rho_i$ gives
\begin{equation}
  G_i^\top G_j
    =\beta_i\beta_jI+g\beta_i\overline R_j
      +g\beta_j\overline R_i^\top
      +g^2\overline R_i^\top\overline R_j.
\end{equation}
The last term has the relative structure of Q/K, while the two linear terms also contain the individual observation addresses. Invariance to a common address translation therefore applies to the Q/K positional kernel; distribution-guided content aggregation generally also uses absolute address information.

\subsection{Physical Addresses and Preprocessing Conventions}
\label{sec:physical_coordinates}

\paragraph{Coordinates and scales.}
In Eq.~\eqref{eq:physical_address}, time is measured in seconds and camera centers retain the input coordinate units (meters for metric poses and reference-baseline-normalized units for RE10K). The scales $s_t,s_o>0$ have the corresponding units, and $s_d>0$ adjusts the direction scale. Static tasks omit the time component of the seven-dimensional address. Robot experiments use camera-to-robot-root extrinsics. When the RGB undistortion, crop/resize, and intrinsic-update chain has not been fully verified, $d$ denotes a nominal ray direction.

\paragraph{Relative addresses.}
The address difference entering the positional operator is
\begin{equation}
  x_j-x_i=
  \bigl[f(\tau_j)-f(\tau_i),\;(o_j-o_i)/s_o,\;s_d(d_j-d_i)\bigr].
\end{equation}
The three terms represent the mapped time difference, camera baseline or motion displacement, and viewing-direction difference. With $f(\tau)=\tau$, the first term recovers the actual time interval scaled by $s_t$. If two unit viewing directions form an angle $\theta$, then $\|d_j-d_i\|_2=2\sin(\theta/2)$. Together, these quantities describe geometric and temporal relationships between observations.

\paragraph{Image preprocessing.}
Pinhole images should first be undistorted, with intrinsics updated consistently after cropping and resizing. If homogeneous pixel coordinates satisfy $\widetilde p'=A_{\rm img}\widetilde p$, choosing $K'=A_{\rm img}K$ yields $K'^{-1}\widetilde p'=K^{-1}\widetilde p$, preserving the back-projected viewing direction of corresponding pixels. Other camera models use their corresponding calibrated back-projection. When the resolution or tokenizer changes, observation distributions are reconstructed from the new pixel footprints.

All addresses containing time use $f(\tau)=2\tanh(\tau/2)$, which is approximately linear near small offsets and limits phase growth at long offsets. The relative-position property applies to the mapped addresses.

\subsection{Numerical Implementation and Complexity}
\label{sec:quadrature_implementation}

\paragraph{Weighted averaging over observation nodes.}
We select $K_i$ deterministic nodes within a token's pixel and temporal footprint, map them to addresses $x_{ik}$, and assign normalized weights $w_{ik}\geq0$ with $\sum_kw_{ik}=1$. The statistics at each frequency are computed directly as
\begin{equation}
  \begin{aligned}
    \widehat c_{i,m}&=\sum_{k=1}^{K_i}w_{ik}\cos(\omega_m^\top x_{ik}),\\
    \widehat s_{i,m}&=\sum_{k=1}^{K_i}w_{ik}\sin(\omega_m^\top x_{ik}).
  \end{aligned}
  \label{eq:finite_frequency_statistics}
\end{equation}
Substituting this pair of statistics into Eq.~\eqref{eq:expected_rotation} gives the numerical positional operator, which is reused across all attention interactions involving that token.

\paragraph{Spatiotemporal quadrature nodes.}
Visual tokens use $3\times3$ spatial nodes on $\{1/6,3/6,5/6\}^2$, with weights
\begin{equation}
  w_{ab}\propto\exp\!\left(
  -\frac{\lVert p_{ab}-(1/2,1/2)\rVert_2^2}{2\sigma^2}
  \right),\qquad\sigma=0.25
\end{equation}
normalized to form the observation weights. For temporal quadrature, we assign each latent token a nominal bin of at most four source frames according to the tokenizer's temporal compression stride. This bin specifies our observation nodes rather than the full receptive field of the causal VAE. Combined with nine spatial nodes per frame, this gives at most 36 nodes per token. Table~\ref{tab:method_configurations} lists the frequency bands, scales, and V/O settings for each task.

\paragraph{Computation and storage.}
For $N$ tokens, $M=d_h/2$ frequencies, and a fixed address dimension, computing the statistics costs $\mathcal O(M\sum_iK_i)$ and requires $\mathcal O(NM)$ storage. Q/K encoding and V/O transformations each cost $\mathcal O(Nd_h)$, while the dominant attention term remains $\mathcal O(N^2d_h)$. Section~\ref{sec:efficiency} reports measured latency.

\section{Complete Numerical Results}
\label{sec:efficiency_details}

\subsection{Per-Case Latency}

Table~\ref{tab:latency_detail} reports raw median latencies for seven cases. Effective token counts already include the batch dimension; each entry lists forward time above forward+backward time. RayRoPE fails on T3 and T5 because the reference implementation does not accept the corresponding multiframe shapes, so these entries remain missing. Each case uses 10 warm-up runs and 30 timed measurements.

\begin{table}[ht]
\centering
\papertablestyle
\setlength{\tabcolsep}{3pt}
\caption{Median per-case model-block latency in milliseconds. Each cell lists forward time above forward+backward time; bold marks the minimum in each case and pass.}

\begin{tabular}{@{}clrrrrrr>{\columncolor{oursrow}}r>{\columncolor{oursrow}}r@{}}
\toprule
Case & Image size & B & T & Tokens & \shortstack{Standard\\RoPE} & P-RoPE & RayRoPE & \shortstack{POD-RoPE\\(Q/K-only)} & POD-RoPE\\
\midrule
R0 & $128\times160$ & 1 & 1 & 240 & \shortstack{4.907\\\textbf{10.552}} & \shortstack{13.395\\39.429} & \shortstack{41.789\\102.995} & \shortstack{\textbf{4.717}\\12.028} & \shortstack{7.234\\16.496}\\[3pt]
R1 & $256\times320$ & 1 & 1 & 960 & \shortstack{5.390\\\textbf{11.581}} & \shortstack{13.812\\38.569} & \shortstack{39.319\\100.942} & \shortstack{\textbf{4.577}\\11.737} & \shortstack{6.537\\17.343}\\[3pt]
R2 & $336\times448$ & 1 & 1 & 1764 & \shortstack{\textbf{6.037}\\\textbf{16.795}} & \shortstack{14.019\\38.312} & \shortstack{41.616\\105.185} & \shortstack{6.624\\17.461} & \shortstack{7.736\\19.718}\\[3pt]
B2 & $256\times320$ & 2 & 1 & 1920 & \shortstack{\textbf{5.454}\\\textbf{15.653}} & \shortstack{13.340\\37.914} & \shortstack{41.385\\106.076} & \shortstack{6.087\\16.423} & \shortstack{7.270\\19.861}\\[3pt]
B4 & $256\times320$ & 4 & 1 & 3840 & \shortstack{\textbf{9.099}\\\textbf{27.395}} & \shortstack{16.545\\47.976} & \shortstack{48.219\\107.367} & \shortstack{9.938\\28.333} & \shortstack{11.840\\32.045}\\[3pt]
T3 & $256\times320$ & 1 & 3 & 2880 & \shortstack{\textbf{9.683}\\\textbf{28.712}} & \shortstack{15.705\\45.628} & \shortstack{--\\--} & \shortstack{10.394\\29.611} & \shortstack{11.894\\32.890}\\[3pt]
T5 & $256\times320$ & 1 & 5 & 4800 & \shortstack{\textbf{16.942}\\\textbf{55.506}} & \shortstack{24.996\\75.113} & \shortstack{--\\--} & \shortstack{17.874\\56.632} & \shortstack{20.176\\62.116}\\[3pt]
\bottomrule
\end{tabular}
\label{tab:latency_detail}
\end{table}

\subsection{Complete RE10K Metrics}

Figure~\ref{fig:efficiency_comparison}(b) in the main text uses forward-plus-backward latency from the matched RE10K/LVSM timing protocol as its speed axis. Table~\ref{tab:re10k} supplies PSNR, SSIM, and LPIPS under the same 80k training protocol. These timings are reported separately from the reconstruction-block microbenchmarks above.

\begin{table}[htbp]
\centering
\papertablestyle
\caption{LVSM novel-view synthesis on RE10K with 80k steps and $n=6474$. Bold denotes the best result; pale blue identifies POD-RoPE variants.}

\begin{tabular}{@{}lrrr@{}}
\toprule
Method & PSNR$\uparrow$ & SSIM$\uparrow$ & LPIPS$\downarrow$\\
\midrule
Pl\"ucker Raymap & 23.442 & 0.7490 & 0.1390\\
GTA & 24.308 & 0.7790 & 0.1260\\
P-RoPE & 24.503 & 0.7838 & 0.1118\\
RayRoPE & \textbf{26.055} & \textbf{0.8312} & \textbf{0.0842}\\
\rowcolor{oursrow}\qkmethod & 24.825 & 0.7970 & 0.1020\\
\rowcolor{oursrow}\method & 25.223 & 0.8061 & 0.0994\\
\bottomrule
\end{tabular}
\label{tab:re10k}
\end{table}

\section{Policy Results on All RoboTwin 2.0 Tasks}
\label{sec:fastwam_details}
Table~\ref{tab:fastwam50_tasks} reports all 50 tasks under the same training budget and evaluation settings. Our method uses the 30k checkpoint; implementation details are given in Appendix~\ref{sec:fastwam_setup}. Both groups evaluate 100 episodes per task and condition, with equal weighting across tasks.

\papertablestyle
\begin{longtable}{@{}lrr>{\columncolor{oursrow}}r>{\columncolor{oursrow}}r@{}}
\caption{Fast-WAM success rates (\%) on all 50 RoboTwin 2.0 tasks, using the methods and settings of Table~\ref{tab:fastwam50}.}\label{tab:fastwam50_tasks}\\
\toprule
\multirow{2}{*}{Task} & \multicolumn{2}{c}{Baseline} & \multicolumn{2}{c}{Ours}\\
\cmidrule(lr){2-3}\cmidrule(lr){4-5}
 & Clean & Rand. & Clean & Rand.\\
\midrule
\endfirsthead
\toprule
\multirow{2}{*}{Task} & \multicolumn{2}{c}{Baseline} & \multicolumn{2}{c}{Ours}\\
\cmidrule(lr){2-3}\cmidrule(lr){4-5}
 & Clean & Rand. & Clean & Rand.\\
\midrule
\endhead
\midrule
\multicolumn{5}{r}{Continued on next page}\\
\endfoot
\bottomrule
\endlastfoot
adjust\_bottle & 100 & 100 & 100 & 100\\
beat\_block\_hammer & 99 & 97 & 100 & 98\\
blocks\_ranking\_rgb & 100 & 100 & 100 & 100\\
blocks\_ranking\_size & 94 & 98 & 95 & 94\\
click\_alarmclock & 100 & 100 & 100 & 100\\
click\_bell & 100 & 100 & 100 & 100\\
dump\_bin\_bigbin & 97 & 96 & 96 & 96\\
grab\_roller & 100 & 100 & 100 & 100\\
handover\_block & 95 & 81 & 92 & 82\\
handover\_mic & 99 & 100 & 99 & 99\\
hanging\_mug & 58 & 62 & 55 & 63\\
lift\_pot & 100 & 100 & 100 & 99\\
move\_can\_pot & 90 & 88 & 94 & 94\\
move\_pillbottle\_pad & 100 & 99 & 100 & 99\\
move\_playingcard\_away & 100 & 100 & 100 & 100\\
move\_stapler\_pad & 77 & 64 & 80 & 81\\
open\_laptop & 98 & 100 & 96 & 99\\
open\_microwave & 62 & 45 & 70 & 59\\
pick\_diverse\_bottles & 80 & 85 & 83 & 85\\
pick\_dual\_bottles & 100 & 96 & 100 & 97\\
place\_a2b\_left & 95 & 93 & 99 & 98\\
place\_a2b\_right & 93 & 99 & 99 & 98\\
place\_bread\_basket & 91 & 93 & 95 & 95\\
place\_bread\_skillet & 90 & 93 & 93 & 93\\
place\_burger\_fries & 96 & 99 & 97 & 97\\
place\_can\_basket & 71 & 69 & 68 & 70\\
place\_cans\_plasticbox & 99 & 96 & 95 & 98\\
place\_container\_plate & 96 & 100 & 99 & 100\\
place\_dual\_shoes & 94 & 88 & 93 & 88\\
place\_empty\_cup & 100 & 100 & 100 & 100\\
place\_fan & 96 & 96 & 96 & 97\\
place\_mouse\_pad & 83 & 89 & 90 & 90\\
place\_object\_basket & 89 & 88 & 90 & 86\\
place\_object\_scale & 90 & 97 & 97 & 95\\
place\_object\_stand & 90 & 94 & 92 & 95\\
place\_phone\_stand & 97 & 99 & 95 & 98\\
place\_shoe & 96 & 99 & 96 & 97\\
press\_stapler & 90 & 97 & 97 & 94\\
put\_bottles\_dustbin & 95 & 90 & 90 & 90\\
put\_object\_cabinet & 94 & 89 & 92 & 92\\
rotate\_qrcode & 93 & 89 & 91 & 88\\
scan\_object & 89 & 92 & 93 & 93\\
shake\_bottle & 100 & 100 & 100 & 100\\
shake\_bottle\_horizontally & 100 & 100 & 100 & 100\\
stack\_blocks\_three & 95 & 97 & 96 & 96\\
stack\_blocks\_two & 100 & 100 & 99 & 100\\
stack\_bowls\_three & 80 & 81 & 84 & 82\\
stack\_bowls\_two & 92 & 98 & 97 & 97\\
stamp\_seal & 90 & 94 & 88 & 91\\
turn\_switch & 61 & 59 & 75 & 69\\
\midrule
Average (50 tasks) & 91.88 & 91.78 & \textbf{93.12} & \textbf{92.64}\\
$\Delta$ (pp) & -- & -- & +1.24 & +0.86\\
\end{longtable}
\normalsize

\section{Reconstruction under Input Changes}
\label{sec:embodied_input_changes}

\paragraph{Shared protocol.}
We reuse the nine 2500-step checkpoints from Sharpa North, RoboTwin, and AGIBot, changing image processing or historical-frame sampling with fixed model weights and identical samples across methods. Every condition processes actual images, updates intrinsics $K$ and actual times, and re-encodes them with the frozen Wan2.1 VAE, yielding 50,700 predictions. Evaluation includes latent MSE and PSNR, SSIM, and LPIPS against both context-consistent VAE-decoded targets and original RGB. This experiment changes the reference inputs, unlike conditional temporal extrapolation, which fixes the references and moves the future target.

\begin{table}[!htbp]
\centering\papertablestyle
\caption{Input-change conditions. Each transformation updates effective intrinsics and actual times consistently.}
\label{tab:embodied_shift_conditions}
\begin{tabular}{@{}>{\raggedright\arraybackslash}p{3.4cm}>{\raggedright\arraybackslash}p{10.0cm}@{}}
\toprule
Condition & Processing\\
\midrule
base & Re-encode under this suite's baseline; target times and scoring differ from normal reconstruction and temporal $h0$.\\
res336 & Resize from $256\times320$ to $336\times448$, changing pixel count and aspect ratio.\\
crop70; fov\_mixed\_a/b & Center-crop to 70\% and resize, or apply camera-specific crop ratios; use only the available image area.\\
native\_fov70/90/130 & AGIBot only: rectify raw fisheye images to virtual pinhole views with 70/90/130-degree FOV.\\
stride5/8; irregular & Sample history at stride 5/8 or irregular intervals; use sufficiently long clips.\\
res336\_stride5/8 & Combine res336 with the corresponding history sampling stride.\\
\bottomrule
\end{tabular}
\end{table}

Table~\ref{tab:embodied_shift_details} reports the PSNR difference \method$-$P-RoPE under the same condition, rather than changes relative to each method's own base condition. Metrics are first averaged across tasks within each trajectory and then across trajectories; RoboTwin aggregates three target directions. The 95\% intervals use trajectory-level paired bootstrap with 2000 resamples and reflect uncertainty across evaluation trajectories at fixed checkpoints. Sharpa North uses 50 trajectories. RoboTwin uses 166 sufficiently long trajectories for sampling conditions and all 271 for the remaining image conditions. AGIBot uses 20 trajectories. Each condition is compared with base on the same subset of trajectories, anchors, and target tasks.

\begin{table}[ht]
\centering\papertablestyle
\caption{PSNR differences for \method$-$P-RoPE under all transfer conditions, with trajectory-level paired 95\% confidence intervals. RoboTwin aggregates three target directions. Each transfer condition is paired with base on the same eligible subset. Dashes indicate undefined conditions.}
\label{tab:embodied_shift_details}

\begin{tabular}{@{}lrrr@{}}
\toprule
Condition & Sharpa North & RoboTwin & AGIBot\\
\midrule
base & +0.751 [+0.663, +0.837] & -0.594 [-0.687, -0.505] & +0.406 [+0.219, +0.567]\\
res336 & +4.174 [+3.911, +4.453] & +3.328 [+3.101, +3.560] & +2.456 [+2.058, +2.803]\\
crop70 & +1.898 [+1.775, +2.025] & +0.568 [+0.418, +0.725] & +1.442 [+1.258, +1.625]\\
fov\_mixed\_a & +2.507 [+2.392, +2.624] & +0.319 [+0.191, +0.450] & +1.300 [+1.078, +1.534]\\
fov\_mixed\_b & +1.388 [+1.274, +1.503] & +0.584 [+0.448, +0.725] & +1.580 [+1.365, +1.771]\\
stride5 & -0.671 [-0.846, -0.509] & -0.582 [-0.700, -0.476] & +0.393 [+0.220, +0.551]\\
stride8 & -0.694 [-0.871, -0.529] & -0.590 [-0.705, -0.486] & +0.386 [+0.211, +0.544]\\
irregular & -0.617 [-0.785, -0.450] & -0.572 [-0.689, -0.467] & +0.400 [+0.227, +0.556]\\
res336\_stride5 & +3.718 [+3.444, +4.003] & +2.750 [+2.495, +3.015] & +2.419 [+2.026, +2.749]\\
res336\_stride8 & +3.359 [+3.097, +3.636] & +2.368 [+2.142, +2.604] & +2.380 [+1.986, +2.708]\\
native\_fov70 & --- & --- & +0.779 [+0.550, +1.039]\\
native\_fov90 & --- & --- & +1.142 [+0.973, +1.316]\\
native\_fov130 & --- & --- & +1.347 [+1.093, +1.542]\\
\bottomrule
\end{tabular}
\end{table}

\paragraph{Results.}
Under res336, PSNR, SSIM, and LPIPS improve over P-RoPE on all three datasets. Scoring against original RGB still gives PSNR differences of $+4.160$, $+3.321$, and $+2.473$ dB on Sharpa North, RoboTwin, and AGIBot, respectively. PSNR-difference intervals for cropping and mixed FOV are all above zero, but LPIPS is slightly worse on RoboTwin and does not improve under AGIBot's native 70-degree FOV. When only the sampling interval changes, POD-RoPE has lower PSNR than P-RoPE under stride5/8 and irregular sampling on both Sharpa North and RoboTwin. When combined with size changes, the PSNR differences become positive.

\section{Experimental Settings and Supplementary Analysis}
\label{sec:experiment_setup_details}

\subsection{Method Integration and Training Budgets}
\paragraph{Policy integration and parameter overhead.}
POD-RoPE integrates into existing attention layers or through lightweight residual attention modules, depending on the policy architecture. For $\pi_0$ and ACT, positional and content transformations operate within the existing visual attention, requiring no additional adapters or trainable parameters. Fast-WAM in simulation and RDT use residual attention modules of width 192, adding 3.154M and 0.887M trainable parameters, respectively, equivalent to approximately 0.052\% and 0.072\% of their respective baselines' trainable parameter counts, excluding frozen encoders. The reported evaluations show higher average task success for all four policy architectures, supporting the practical use of POD-RoPE with zero or small parameter increments. The robot experiments evaluate the complete integrations, while the controlled ablations in Table~\ref{tab:component_ablation} assess the contributions of distribution encoding and V/O transformations.

\begin{table}[t]
\centering
\papertablestyle
\setlength{\tabcolsep}{3.8pt}
\caption{Experimental integration configurations for POD-RoPE. Q/K denotes distribution-based positional encoding and V/O denotes content transformations. L2 and RMSNorm precede the positional operator; input LN precedes projection.}
\label{tab:method_configurations}
\begin{tabular}{@{}>{\raggedright\arraybackslash}p{3.5cm}lcccc@{}}
\toprule
Model / variant & Address & Bands & Scale & Norm. & V/O\\
\midrule
Reconstruction & Space-time & $\mathcal B$ & S & L2 & Yes\\
$\pi_0$ & Space-time & $\mathcal B$ & S & None & Yes\\
RDT & Space-time & $\mathcal B$ & S & Input LN & Yes\\
ACT & Space-time & $\mathcal B$ & S & None & Yes\\
Fast-WAM (sim., vision) & Space-time & $\mathcal B$ & S & Input LN & Yes\\
Fast-WAM (sim., action) & Sequence & A & A & Input LN & Yes\\
Fast-WAM (real, vision) & Space-time & $\mathcal B$ & S & Input LN & Yes\\
Fast-WAM (real, action) & Sequence & A & A & Input LN & Yes\\
UniMatch (Q/K-only) & Spatial & L & U & L2 & No\\
UniMatch (Q/K+V/O) & Spatial & L & U & L2 & Yes\\
LVSM (Q/K-only) & Spatial & H & R & RMSNorm & No\\
LVSM (Q/K+V/O) & Spatial & H & R & RMSNorm & Yes\\
\bottomrule
\end{tabular}
\end{table}

In Table~\ref{tab:method_configurations}, spatiotemporal addresses are $[f(\tau),\bar o,\bar d]$ and spatial addresses omit time, with $f(\tau)=2\tanh(\tau/2)$. S uses $s_t=0.125\,\mathrm{s}$, $s_o=0.7504372976$, and $s_d=103.01593647$; R uses the same numerical spatial scales, with camera centers expressed in reference-baseline-normalized units; S retains the units of its input camera centers. U directly uses $\bar o=4.2127334\,o$ and $\bar d=245.97024\,d$, retaining the DeMoN input units for camera-center coordinates. S/R divide camera centers by $s_o$, while U multiplies them by the coefficient above.

The six-band set is $\mathcal B=\{0.0625,0.125,0.25,0.5,1,2\}$, assigned cyclically to feature pairs. Let $P=d_h/2$ and $r=0,\ldots,P-1$. L uses frequency magnitudes $10{,}000^{-r/(P-1)}$; H further multiplies them by $1+h/H$, where $h=0,\ldots,H-1$ is the head index. Fast-WAM's action branch A uses $10{,}000^{-r/P}$ and maps the $n$th action and its nine subinterval nodes $\delta_k$ to $2\tanh((n+\delta_k)/64)$, corresponding to a reference length of 32. This encodes action-sequence positions. Each visual distribution uses nine spatial nodes per source frame with $\sigma=0.25$. All configurations with V/O enabled fix $g=1$.

\paragraph{Complete frequency vectors.}
The address coordinates are ordered as time, camera-center $xyz$, and unit-ray $xyz$; static addresses omit time. For address dimension $D$, coordinate index $a=1,\ldots,D$, and zero-based head and feature-pair indices $h,r$, we construct
\begin{equation}
\begin{aligned}
u_{h,r,a}&=\sin\!\big((hP+r+1)a\sqrt{2}\big)+\cos\!\big((hP+r+1)a\sqrt{3}\big),\\
\omega_{h,r}&=b_{h,r}\,u_{h,r}/\|u_{h,r}\|_2,
\end{aligned}
\label{eq:complete_frequency_vectors}
\end{equation}
where $b_{h,r}$ is the magnitude specified by the frequency configuration above; for $\mathcal B$, $b_{h,r}=\mathcal B_{r\bmod 6}$ with zero-based indexing. The $\pi_0$ integration shares directions across heads, replacing $hP+r$ with $r$.

\paragraph{Temporal support and weights.}
In reconstruction, $t_0$ is the minimum timestamp over all reference and target support nodes in the current sample, and the temporal coordinate is $2\tanh((t-t_0)/(2s_t))$, with fixed $s_t=0.125\,\mathrm{s}$. The first latent temporal plane uses raw frame 0; plane $l\geq1$ uses raw frames $4l-3,\ldots,4l$ with equal temporal weights. Each frame contributes Gaussian-9 spatial nodes. Joint weights are the products of temporal and spatial weights, normalized over the token's support.

\paragraph{Spatial scale calibration.}
On RE10K, $s_o=0.7504372976$ is the median baseline length of selected training-view pairs, and $s_d=103.01593647$ is the reciprocal of the median within-token Gaussian-weighted angular RMS of the rays. Angles are measured relative to the normalized weighted mean ray, using the training preprocessing and spatial nodes. RE10K translations are normalized by the reference-view baseline, so $s_o$ is expressed in normalized baseline units. Other settings with the same numerical scales in Table~\ref{tab:method_configurations} reuse these constants.

\begin{table}[!htbp]
\centering\papertablestyle
\caption{Training budgets for visual experiments. Backbone entries list layers / width / heads / FFN width. The ablation and main reconstruction experiments use different backbones.}
\label{tab:training_budgets}
\begin{tabular}{@{}llrr@{}}
\toprule
Experiment & Backbone & Global batch & Updates\\
\midrule
Reconstruction & 24 / 1152 / 8 / 4608 & 2,048 & 2,500\\
Component ablation & 6 / 1152 / 8 / 1024 & 512 & 2,500\\
UniMatch & 6 blocks / 144 ch. / 1 head & 16 & 100,000\\
RE10K/LVSM & 6 / 1152 / 8 / 1024 & 8 & 80,000\\
\bottomrule
\end{tabular}
\end{table}

\paragraph{Attention operation order.}
\label{sec:attention_order}
Each model first projects Q/K, then applies the positional operator $M$, and finally computes scaled dot products. L2 normalization in Table~\ref{tab:method_configurations} is followed by multiplication by $\sqrt{d_h}$; both L2 and RMSNorm precede $M$, while input LN precedes projection. ACT and $\pi_0$ do not separately normalize Q/K in the integration. The full method aggregates content as $G_jv_j$ and applies $G_i^\top$. Table~\ref{tab:method_configurations} summarizes addresses, frequency bands, and scales for all integrations.

\subsection{Embodied Reconstruction and Conditional Temporal Extrapolation}
\begin{figure}[!htbp]
\centering
\paperfigure[width=\textwidth]{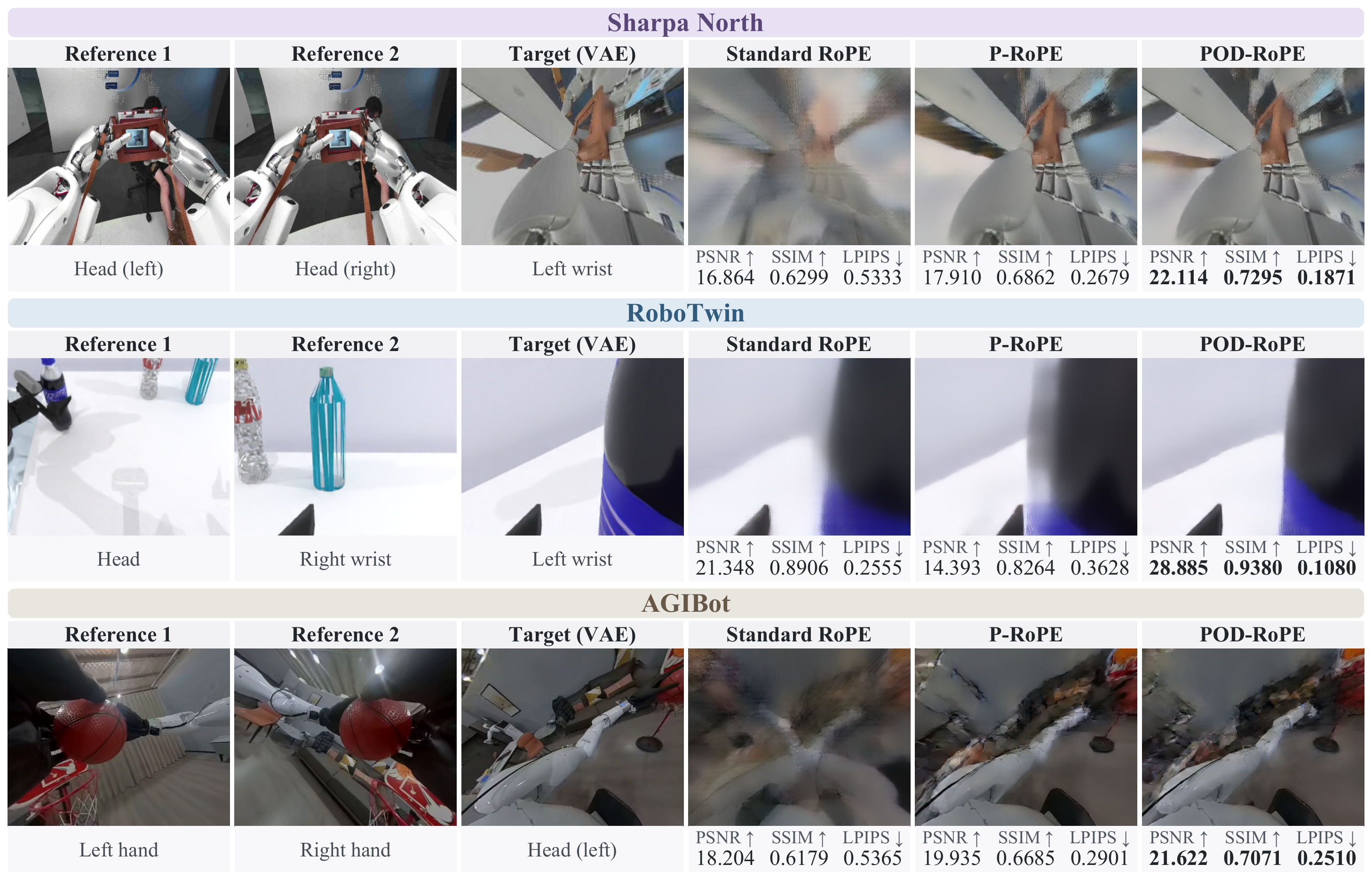}
\caption{Synchronized cross-view reconstruction examples. Rows show Sharpa North, RoboTwin, and AGIBot from top to bottom. Each model reconstructs a third camera view from two simultaneous reference observations. Columns show Reference 1, Reference 2, the VAE-decoded target, and reconstructions from Standard RoPE, P-RoPE, and POD-RoPE. Metrics below the images are PSNR (dB), SSIM, and LPIPS, respectively; higher is better for the first two and lower is better for the last. Bold marks the best value for each example. Appendix~\ref{sec:reconstruction_evaluation} specifies the scoring windows.}
\label{fig:embodied_reconstruction}
\end{figure}

\paragraph{Models and data.}
On each of the three datasets, we separately train Standard RoPE (axial 3D), P-RoPE~\citep{li2025cameras}, and a fixed \method configuration (Gaussian-9, bounded temporal mapping, and Q/K+V/O). Table~\ref{tab:training_budgets} gives the training budgets. Inputs are $256\times320$, Wan2.1 VAE is frozen, and the model receives a single latent plane. Training uses BF16, AdamW, a learning rate of $2\times10^{-4}$, and shared splits. Standard RoPE and \method each have 382,409,856 parameters; P-RoPE has 382,416,768.

\begin{table}[!htbp]
\centering\papertablestyle
\caption{Data and tasks for embodied reconstruction. Sample counts give the ordinary reconstruction / $h16$ evaluation totals per method. Splits are made by trajectory. Models are trained independently on each dataset, and all three methods share the split within a dataset.}
\label{tab:embodied_protocols}

\begin{tabular}{@{}lcc>{\raggedright\arraybackslash}p{4.5cm}@{}}
\toprule
Dataset & \shortstack{Train / val.\\trajectories} & \shortstack{Eval. samples\\(normal / $h16$)} & Target task\\
\midrule
Sharpa North & 281 / 50 & 200 / 200 & Head--wrist cross-view\\
RoboTwin & 2,212 / 271 & 1,800 / 1,728 & Head and both wrists as targets\\
AGIBot & 180 / 20 & 80 / 80 & Head--hand cross-view\\
\bottomrule
\end{tabular}
\end{table}

The datasets are, in order, our collected Sharpa North head-stereo and dual-wrist camera data, RoboTwin, and the public AgiBot World dataset (AGIBot)~\citep{bu2025agibot}. Each task reconstructs a third view from two references; Table~\ref{tab:embodied_protocols} gives camera roles and dataset sizes.

\paragraph{Scoring and temporal protocol.}
\label{sec:reconstruction_evaluation}
Metrics are averaged across tasks within each trajectory and then across trajectories. We compute 95\% intervals with trajectory-level paired bootstrap using 2000 resamples. PSNR, SSIM, and LPIPS use context-consistent VAE-decoded targets as references; latent MSE is reported separately. Conditional temporal extrapolation fixes the two reference observations, camera roles, and target trajectory, and uses actual future targets at $h\in\{0,1,2,4,8,16\}$. One step spans four original synchronized sampling positions. Mean delays at $h16$ are 2.151, 4.267, and 2.133 seconds on Sharpa North, RoboTwin, and AGIBot, respectively; Table~\ref{tab:embodied_protocols} gives sample counts. The predictor receives target camera geometry and time but does not access the future target latent. Target history is used only for shared contextual decoding and scoring. This protocol therefore evaluates conditional reconstruction with known target geometry. Images show the last frame; metrics cover one or four frames according to the latent position.

\begin{figure}[!htbp]
\centering
\paperfigure[width=\textwidth]{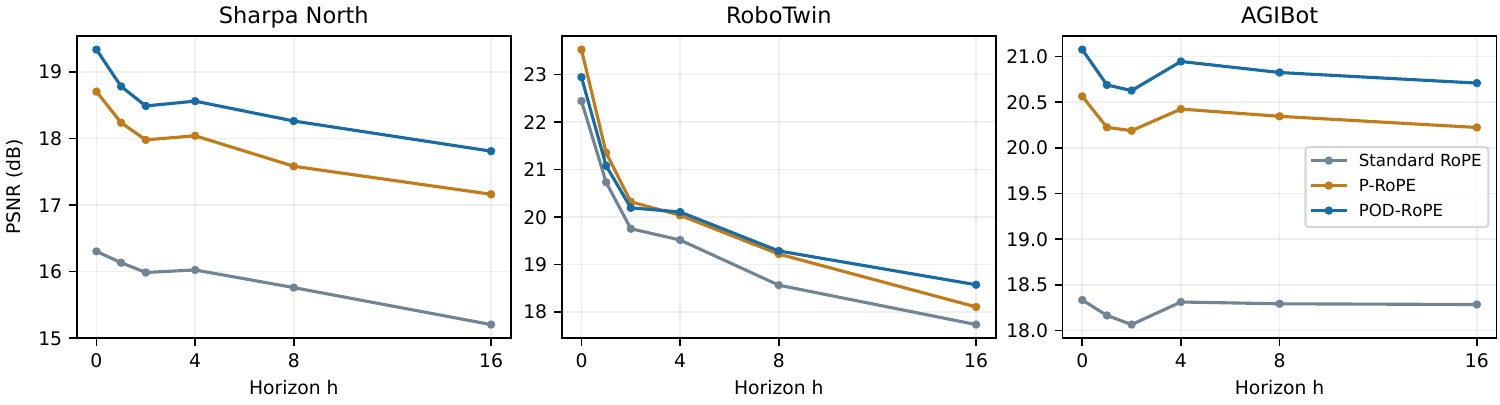}
\caption{Temporal extrapolation to actual targets with fixed references, showing Sharpa North, RoboTwin, and AGIBot from left to right. All points use the same step-2500 checkpoint. Evaluated horizons are $h=0,1,2,4,8,16$; connecting lines are visual guides only. The same $h$ corresponds to different actual delays across datasets. Curves show trajectory-balanced means; RoboTwin aggregates three target directions, as in the main table. Numerical results at $h16$ are reported in Table~\ref{tab:embodied_h16}.}
\label{fig:embodied_time}
\end{figure}

\paragraph{Results across temporal horizons.}
\label{sec:temporal_horizon_analysis}
Figure~\ref{fig:embodied_time} provides all temporal horizons. On RoboTwin, our method is 0.581 dB below P-RoPE at $h0$ in this suite; difference intervals at $h4/h8$ cross zero, while all four metrics lead at $h16$. LPIPS does not improve on AGIBot. Temporal changes should be compared against $h0$ within this suite.

\subsection{Single-Task Evaluation across Policy Architectures}
\label{sec:robotwin_single_details}

On 15 RoboTwin 2.0 tasks, we follow each model's corresponding baseline training recipe and evaluation settings, independently training each model for each task. Each task uses 50 Clean demonstrations. ACT uses 40 training and 10 validation demonstrations; Table~\ref{tab:policy_training} gives the remaining recipes. Clean/Randomized correspond to Easy/Hard. Table~\ref{tab:robotwin_single_all_tasks} in the main text reports complete results.

\begin{table}[!htbp]
\centering\papertablestyle
\caption{Single-task policy training and checkpoint selection. Both evaluation conditions use the same checkpoint.}
\label{tab:policy_training}
\begin{tabular}{@{}lrl>{\raggedright\arraybackslash}p{2.8cm}>{\raggedright\arraybackslash}p{3.4cm}@{}}
\toprule
Model & Global batch & Budget & \shortstack[l]{Training /\\checkpoint} & Execution\\
\midrule
$\pi_0$ & 32 & 30,000 updates & 4 GPUs, FSDP; final EMA & 50 actions per call\\
RDT & 64 & 10,000 updates & 4 GPUs; final checkpoint & Native qpos\\
ACT & 8 & 6,000 epochs & 1 GPU; best validation & Chunk 50; temporal aggregation\\
\bottomrule
\end{tabular}
\end{table}

\paragraph{Evaluation.}
For each task and environment, 100 scenes are selected using the native expert feasibility check. We use native qpos execution and task-success criteria. Clean uses seen instructions and Randomized uses unseen instructions. Success rates are averaged equally across the 15 tasks.

\subsection{Fast-WAM Simulation Evaluation across All Tasks}
\label{sec:fastwam_setup}
Joint training on 50 tasks uses 27,500 trajectories, AdamW, a learning rate of $10^{-4}$, and a 30k-step budget. Data splits, image decoding, per-sample language supervision, and action execution follow the baseline pipeline. Each task runs 100 episodes under each of Clean/Randomized conditions, totaling 10,000 episodes. Results are first averaged equally over tasks, then equally over the two conditions. Both groups use identical training steps and evaluation episode counts; Appendix~\ref{sec:fastwam_details} gives per-task results.

\subsection{Visual Geometry and Efficiency Evaluation}
\paragraph{UniMatch/DeMoN.}
Five positional mechanisms are trained from scratch in UniMatch's~\citep{xu2023unimatch} six Transformer blocks with single-head cross-image attention, using the combined DeMoN~\citep{ummenhofer2017demon} protocol, $448\times576$ crops, eight GPUs, and a learning rate of $4\times10^{-4}$. Table~\ref{tab:training_budgets} gives the budgets. Q/K-only uses six-dimensional addresses comprising camera origins and rays in world coordinates, Gaussian-9 with $\sigma=0.25$, and scales calibrated on the training set. Q/K are normalized to $\sqrt{144}$ before the positional operator. Both variants use the same L frequency configuration; the full method additionally enables V/O with $g=1$. RGBD and Scenes11 are evaluated with \texttt{validate\_demon} at the same 100k checkpoint. We implement P-RoPE and RayRoPE~\citep{li2025cameras,wu2026rayrope} from their published formulations and compare them under the same protocol.

\paragraph{RE10K.}
On RE10K~\citep{zhou2018stereo}, we use the small LVSM~\citep{jin2025lvsm} configuration with $256\times256$ inputs, Q/K normalization, CamRay, FP16, and eight-GPU training. Both \qkmethod and full \method use Gaussian-9, $\sigma=0.25$, the same H frequency configuration, and calibrated spatial scales. The full method enables V/O with $g=1$, without predicting or sampling scene depth. We report PSNR, SSIM, and LPIPS on a fixed set of 6,474 test samples.

\paragraph{Efficiency microbenchmarks.}
We benchmark the same six-layer reconstruction block on one H20 in BF16 using random calibrated rays and latents. Table~\ref{tab:latency_detail} gives seven configurations, each with 10 warm-up runs and 30 measurements. Relative latency is the median of per-configuration ratios. RayRoPE is restricted to the five executable single-slice configurations. Figure~\ref{fig:efficiency_comparison}(a) fixes $256\times320$, B1, T1 (960 tokens) and includes GTA under the same protocol. The quality--efficiency plot in Figure~\ref{fig:efficiency_comparison}(b) uses a separate matched RE10K/LVSM timing protocol. Absolute latencies from the two settings are not mixed; neither includes the full data-loading, decoding, optimizer, or simulation pipeline.

\subsection{Component Ablation}
\label{sec:component_ablation_setup}
Using the Sharpa North four-camera cache and existing split, we fix the two head-camera inputs and reconstruct the left or right wrist view at the same time. The six groups share sampling and parameter initialization; Table~\ref{tab:training_budgets} gives the backbone, with 46,168,704 parameters. Inputs are $256\times320$, Wan VAE is frozen, and the latent grid and patch size are $1\times32\times40$ and $1\times2\times2$. Training uses two GPUs with batch 256 per GPU, no gradient accumulation, BF16, AdamW, learning rate $2\times10^{-4}$, $\beta=(0.9,0.95)$, $\epsilon=10^{-8}$, weight decay 0.05, gradient clipping at 1.0, and 125 warm-up steps.

The five physical encodings share the frequency bands $\mathcal B$, scale S, tanh temporal mapping, and Q/K normalization in Table~\ref{tab:method_configurations}. Each source frame uses nine Gaussian nodes ($\sigma=0.25$), with temporal nodes following the tokenizer footprint. Both point configurations first compute the weighted center of the same spatiotemporal addresses, then normalize the direction and restore its scale; the time component is mapped before averaging. Distribution Q/K retains the expected rotation over all nodes with $g=0$, and the full method changes only $g$ to 1. The direct-average variant keeps the same distribution Q/K and replaces the V/O operators with $\overline R$ and $\overline R^\top$, respectively, omitting the identity branch. Standard RoPE uses a different axial 3D frequency layout, so component effects are compared within the five physical configurations by varying point versus distribution encoding and enabling V/O. All groups evaluate the same 200 validation samples at the 2500-step checkpoint, score against VAE-decoded targets, and take arithmetic means over samples, unlike the trajectory/task-balanced aggregation in the main table.

\subsection{Real-Robot Evaluation on Sharpa North}
\label{sec:real_robot_setup}

\paragraph{Data and observations.}
Both methods train jointly on bottle opening and straw insertion, with task text distinguishing the operations. The two tasks have 343/1,351 training trajectories and 244,097/197,660 training windows, respectively, and 35/185 validation trajectories with 25,141/32,787 validation windows. Splits are made by collection batch, and training mixes valid windows. Preprocessing excludes 304 straw-insertion trajectories with abnormal mobile-base behavior (37,803 windows). Means and standard deviations are estimated only from deduplicated data rows in the retained training windows, with a standard-deviation floor of 0.001.

Images from the head's left eye and both wrists are first remapped to $256\times320$ and then assembled into a $384\times320$ input: the head image is on top, with two $128\times160$ wrist images side by side below. The head uses batch-specific intrinsics and distortion parameters; the wrists use an approximation based on a historical fisheye model. Both methods share this preprocessing. Camera links are mapped to robot root through URDF forward kinematics; this geometry is not an independently and precisely calibrated world coordinate system. Head timestamps define 33-frame windows. Video uses nine frames at offsets $0,4,\ldots,32$, while actions cover 32 steps. The 67-dimensional state/action comprises 7 dimensions per arm, 22 per hand, 7 for other joints, and 2 for mobile-base velocity. The policy reads only current observation geometry.

\paragraph{Models and training.}
Both methods share the pretrained Fast-WAM backbone, data, and identical width-192, six-head modules in the video and action branches. The baseline uses standard RoPE; \method uses Gaussian-9 ($\sigma=0.25$) for Q/K and V/O with $g=1$. Visual tokens use robot-root addresses, actions encode sequence positions, and the original positional mechanism is retained.

Both groups use 8 nodes with 8 H20 GPUs per node, batch 16 per GPU, global batch 1024, no gradient accumulation, BF16, and ZeRO-1, with gradient checkpointing disabled. AdamW uses a learning rate of $10^{-4}$, $\beta=(0.9,0.95)$, weight decay 0.01, and gradient clipping at 1. A 100k-step cosine schedule includes 1500 warm-up steps; both real-robot evaluations use the 20k checkpoint.

\paragraph{Evaluation and examples.}
For bottle opening, the robot grasps the bottle and unscrews the cap; success requires the cap to be fully unscrewed and separated from the bottle opening. For straw insertion, the robot grasps the straw, aligns it with the target container opening, and inserts it; success requires the straw to remain inside the container after release. Each method is evaluated for 20 trials per task. Only trials that complete all task steps count as successes; incomplete trials count as failures. Success rates divide the number of successes by 20, and the mean weights tasks equally; Table~\ref{tab:real_robot} reports the results.

\end{document}